\documentclass[pre,aps,twocolumn,notitlepage,showpacs,floatfix,superscriptaddress]{revtex4-1}
\usepackage{amsmath}
\usepackage{amssymb}
\usepackage{physics}
\usepackage{bm}
\usepackage{graphicx}
\usepackage{hyperref}
\usepackage[margin=1in]{geometry}
\usepackage[dvipsnames]{xcolor}
\usepackage{algorithm}
\usepackage{algpseudocode}
\usepackage[normalem]{ulem}
\usepackage{quantikz}
\usepackage{array}
\usepackage{tikz}
\usetikzlibrary{positioning, arrows.meta, calc, decorations.pathreplacing,
	decorations.markings, backgrounds, fit}
\usepackage{xcolor}
\usetikzlibrary{positioning, arrows.meta, calc, decorations.pathreplacing,
	decorations.markings, backgrounds, fit}
\definecolor{classicalbg}{RGB}{255,220,180}   
\definecolor{classicalborder}{RGB}{200,120,40}
\definecolor{cvbox}{RGB}{255,200,150}
\definecolor{arrowgray}{RGB}{100,100,100}
\definecolor{panelborder}{RGB}{150,150,150}

\usepackage{colortbl}
\usepackage{booktabs}
\usepackage{amsthm}
\theoremstyle{definition}

\definecolor{lightblue}{rgb}{0.85,0.91,0.95}
\definecolor{darkblue}{rgb}{0.0,0.4,0.65}

\begin{document}
\title{Perturbative Contrastive Physical Learning}

\author{Kyungeun Kim}
\email{kkim@math.ubc.ca}
\affiliation{Department of Mathematics, University of British Columbia, Vancouver, BC Canada}
\author{Amanuel Anteneh}
\affiliation{Department of Physics, University of Virginia, 382 McCormick Rd, Charlottesville, VA 22903, USA}
\author{Israel Klich}
\affiliation{Department of Physics, University of Virginia, 382 McCormick Rd, Charlottesville, VA 22903, USA}
\affiliation{Max Planck Institute for the Physics of Complex Systems, 01187 Dresden, Germany}
\author{Olivier Pfister}
\email{opfister@virginia.edu}
\affiliation{Department of Physics, University of Virginia, 382 McCormick Rd, Charlottesville, VA 22903, USA}
\affiliation{Charles L. Brown Department of Electrical and Computer Engineering, University of Virginia, 351 McCormick Road, Charlottesville, VA 22903, USA}
\author{J. M. Schwarz}
\email{jmschw02@syr.edu}
\affiliation{Department of Physics, Syracuse University, Syracuse, NY 13244, USA}
\affiliation{Indian Creek Farm, Ithaca, NY 14850, USA}

\date{\today}

\begin{abstract}
Responses to perturbations are key to understanding physical systems. The ability to contrast such responses by comparing how a system reacts under slightly different conditions provides a mechanism for learning. Here, we introduce Perturbative Contrastive Physical Learning (PCPL), a general framework in which learning emerges from measurable contrasts between physical states produced by controlled changes to inputs, boundary conditions, parameters, or interpreter functions. PCPL unifies and extends prior approaches: Equilibrium Propagation is rooted in contrasts between free and nudged equilibria in energy-based systems, while Frequency Propagation corresponds to contrasts extracted from sinusoidally driven, frequency-demodulated responses. We show that contrast-driven updates can reflect either local sensitivities or global inverse-problem structure, yet do not require centralized gradient computation. Instead, effective learning geometry emerges implicitly from the system’s own physical response, allowing learning behavior to arise without an external processor or explicit backpropagation. We demonstrate PCPL in two platforms: (i) spring networks that update bond stiffness using measured displacements and forces, and (ii) continuous-variable photonic circuits trained via $x$ quadrature measurements and finite-difference estimates of the Jacobian. Both platforms successfully learn classification tasks. We further show that a continuous-variable photonic circuit can be trained to implement analog multiplication, illustrating a step toward more autonomous physical learning systems. 

\end{abstract}

\maketitle

\section{Introduction}

Learning requires the ability to modify behavior based on experience. In biology, learning is often associated with brains and centralized neural processing~\cite{hebb1949organization,dayan2001theoretical}. Yet organisms without nervous systems, such as slime molds, are also capable of learning~\cite{reid2016slime}. Moreover, nonliving brain-inspired systems, such as convolutional neural networks, also learn~\cite{Goodfellow2016}. These observations suggest that learning may emerge from general principles that can be realized directly in physical processes. With this viewpoint, learning arises from measurable changes in system behavior produced by controlled perturbations of inputs, boundary conditions, or internal parameters~\cite{scellier2017equilibrium,wright2022deep,Stern2023,momeni2025training}. Physical learning, therefore, represents a fundamental shift from brain-inspired learning at the software level to learning that emerges directly from physical processes, with the hardware effectively becoming the software~\cite{laydevant2024hardware}. 

A key physical principle underlying such learning is the response of a system to perturbations. Physical systems reveal their structure through how observable quantities change when inputs, boundary conditions, or internal parameters are varied. The ability to contrast these responses naturally provides a mechanism for learning. If a system can be perturbed in a controlled way and the resulting changes in its measurable outputs can be detected, those contrasts can be used to guide updates to modifiable parameters. Learning then becomes a process of probing, measuring, and remodeling a physical system based on how it responds.

Here, we introduce Perturbative Contrastive Physical Learning (PCPL), a general framework that formalizes this principle. In PCPL, parameter updates are driven by measurable contrasts between physical states generated by controlled perturbations. These perturbations may involve changes to inputs, boundary conditions, internal parameters, or readout mappings, but in all cases learning is derived from the difference between two nearby physical responses. The perspective that physical learning can emerge from contrasts between nearby physical states was crystallized by Equilibrium Propagation (EP), which showed that measurable differences between free and weakly nudged equilibria can generate learning signals equivalent to gradient-based optimization~\cite{scellier2017equilibrium,scellier2021deep}. PCPL builds on this insight by treating contrast itself as the central primitive. And rather than requiring an explicit energy function, equilibrium convergence, or symbolic gradient computation, PCPL treats learning as an experimentally accessible operation: apply a perturbation, measure the contrast in observables, and translate that contrast into a parameter update.

This perspective reveals that several existing physical learning schemes can be understood as special cases of a broader contrast principle. In addition to EP~\cite{scellier2017equilibrium,scellier2021deep}, there is Coupled Learning with its experimental realizations~\cite{stern2021supervised,stern2025physical,dillavou2022demonstration,Altman2024,Dillavou2024PNAS} and Multi-mechanism learning with the contrast encoded in different modalities~\cite{Anisetti2023}. Frequency Propagation extracts contrasts from sinusoidally driven, frequency-demodulated responses~\cite{anisetti2024frequency}. In addition, explicit temporal contrasts between free and nudged states using a non-equilibrium memory kernel also make learning possible~\cite{falk2025temporal}. PCPL generalizes this idea beyond equilibrium settings and energy-based formulations, allowing contrast-driven learning to be implemented in systems with complex dynamics, multiple steady states, or black-box measurement and interpretation layers. By elevating the contrast itself to the central primitive, PCPL provides a unifying framework for learning across physical substrates.

These ideas are not limited to classical platforms. Recent work has shown that contrast-based learning rules can also be implemented in quantum systems, where observables replace classical state variables and measurement outcomes provide the signals needed for adaptation. In particular, quantum generalizations of Equilibrium Propagation demonstrate that contrasts between weakly perturbed quantum steady states can be used to extract local update signals through measurable expectation values and correlations~\cite{scellier2024quantum,wanjura2025quantum}. From the PCPL perspective, such schemes represent a specific instance of a broader contrast principle: learning is driven by observable differences between nearby quantum states. Continuous-variable photonic platforms offer a natural testbed for this view as Gaussian operations, which include quantum squeezing, single-photon gain/loss, interferometric couplings, and homodyne detection, enable controlled perturbations and precise readout of system responses---including at the quantum computing level~\cite{gu2009cluster,menicucci2008oneway,pfister2020cvqc,killoran2019continuous,schuld2019evaluating}. In this way, quantum physical learning appears not as a fundamentally different paradigm, but as a direct extension of contrast-driven adaptation into regimes where information is encoded in quantum states and accessed through measurement. While such quantum realizations are an important and rapidly developing direction, the present work focuses on classical mechanical and optical systems, leaving a detailed exploration of quantum platforms to future studies.

The organization of the manuscript is as follows. First, we introduce the overall framework, then we apply the framework to a classical spring network followed by  several photonic examples. We conclude with a discussion containing a summary and both experimental and theoretical implications. 

\section{Perturbative Contrastive Physical Learning: A Two Mode Framework}

We begin with a dataset $\mathcal{D}$ of paired observations
$(\mathbf{x},\mathbf{z})$, where $\mathbf{x}$ denotes the set of inputs and $\mathbf{z}$ denotes the corresponding set of outputs. In physical learning systems, $\mathbf{z}$ represents a physical readout, such as a 
desired displacement or stress pattern. A \emph{batch} refers to a finite subset of such pairs sampled from $\mathcal{D}$. The mapping from inputs $\mathbf{x}$ to measurable outputs $\mathbf{z}$ may involve an initial physical encoding, or preprocessing, stage $\mathbf{y}=F(\mathbf{x})$ followed by a parameterized physical response $G(\mathbf{y},\bm{\theta})=\mathbf{z}$. Here, $G$ denotes the observable response of the physical system under modifiable parameters $\bm{\theta}$ to enable learning. Throughout this work, learning acts on the response map $G$, while $F$ is treated as fixed. This decomposition allows fixed physical architectures to solve more complex tasks through structured input representations rather than architectural expansion. 

\begin{figure*}
\centering
\includegraphics[height=7cm]{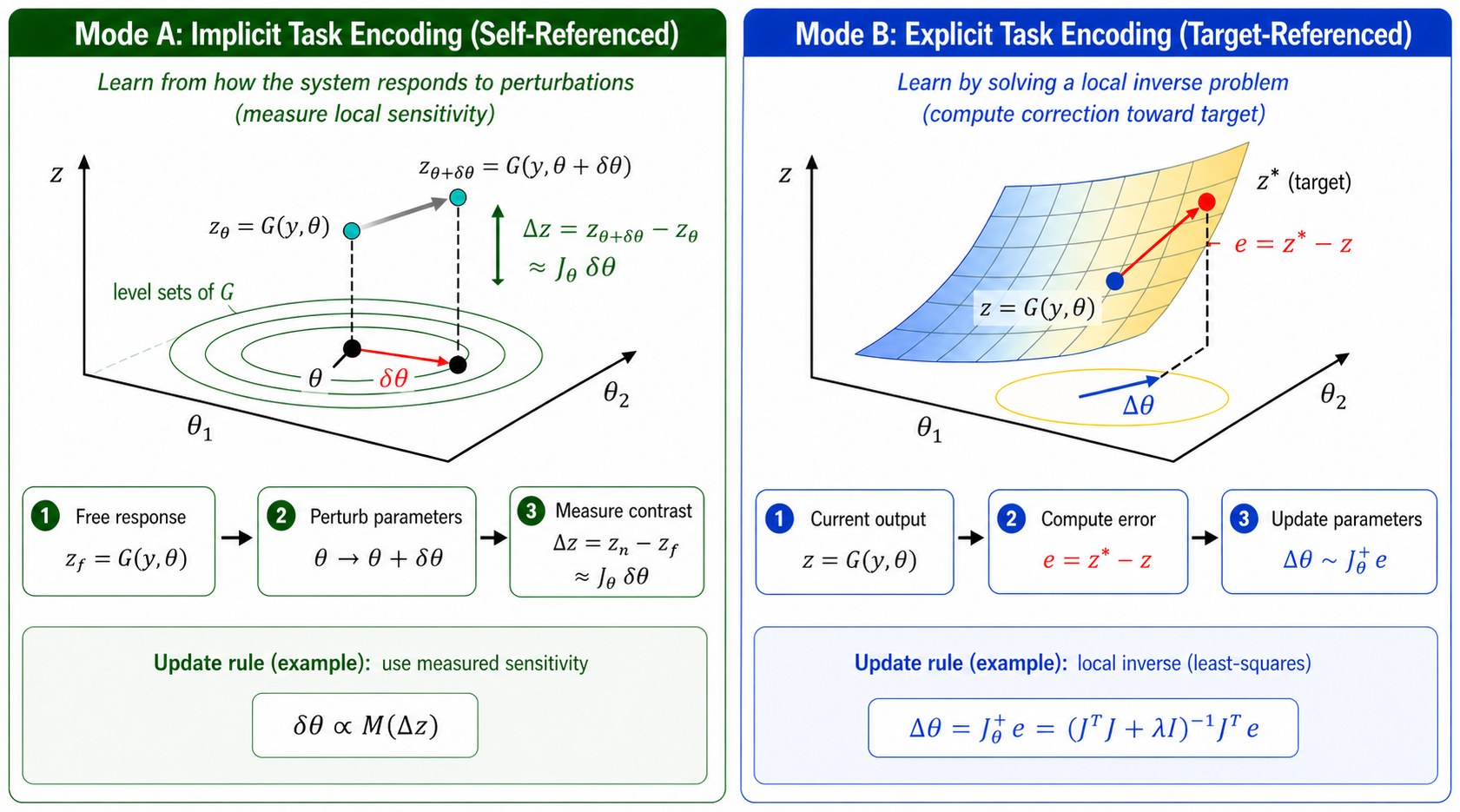}  
\caption{{\it Schematic for PCPL with Mode A (implicit target) and Mode B (explicit target).}}
\end{figure*}
We consider two related learning modes that update parameters
$\bm{\theta}$ of $G$, the response of a physical system, from weak nudging perturbations. In both cases, a batch is sampled, features are computed, a parameter update $\Delta\bm{\theta}$ is applied via $\bm{\theta}'=\bm{\theta}+\Delta\bm{\theta}$. Note that throughout the paper we denote by $\delta\bm{\theta}$ a small imposed probe perturbation used to generate a measurable contrast, and by $\Delta\bm{\theta}$ the parameter update inferred from that contrast. In simple implementations these may be proportional, but they play distinct conceptual roles. To determine $\Delta\bm{\theta}$, the resulting response $G(\mathbf{y},\bm{\theta}')$ is used to form a contrast measurement functional $M$. This functional is task- and implementation-dependent. We will consider learning rules based on measurable contrasts between nearby physical responses via observables $\bm{z}_1$ and $\bm{z}_2$ such that 
\begin{equation}
    \Delta \bm{\theta} \sim M(\bm{z}_1,\bm{z}_2). 
    \end{equation}
In other words, $M$ maps observable contrasts into parameter updates. Specific forms of $M$ will be discussed below. We note that it need not represent an analytically computed gradient; rather, it encodes how physically measurable contrasts can be used for learning.

The distinction between the two modes is the \emph{reference} against which the nudged response is compared: Mode~A contrasts the nudged response against the system's own free response, while Mode~B contrasts it against an externally specified target, often realizable as a physical constraint. Let us now be more specific about the two modes. 

\subsection{Mode A: Self-referenced contrast: Linear response learning}
In Mode A, learning is driven by measurable contrasts between free and weakly perturbed, or nudged, physical responses, enabling parameter updates through local response probing. The system is first allowed to relax in a free phase, producing
\begin{equation}
\mathbf{z}_{\mathrm{f}} \;=\; G(\mathbf{y},\bm{\theta}),
\end{equation}
followed by a nudged phase with $\bm{\theta}'=\bm{\theta}+\delta\bm{\theta}$, yielding
\begin{equation}
\mathbf{z}_{\mathrm{n}} \;=\; G(\mathbf{y},\bm{\theta}+\delta\bm{\theta}).
\end{equation}
For weak nudges, a first-order expansion gives the local \emph{parameter-response map}
\begin{equation}
G(\mathbf{y},\bm{\theta}+\delta\bm{\theta})
\;\approx\;
G(\mathbf{y},\bm{\theta})
\;+\;
J_{\theta}(\mathbf{y},\bm{\theta})\,\delta\bm{\theta},
\qquad
J_{\theta} \equiv \frac{\partial G}{\partial \bm{\theta}}.
\label{eq:linresp}
\end{equation}

Thus, the contrast between phases is
\begin{equation}
\Delta \mathbf{z}
\;\equiv\;
\mathbf{z}_{\mathrm{n}}-\mathbf{z}_{\mathrm{f}}
\;\approx\;
J_{\theta}\,\delta\bm{\theta},
\label{eq:deltaG}
\end{equation}
which is directly accessible as a physical measurement. 

The Mode A learning signal depends on the phase contrast relative to target $\mathbf{z}^*$ via $M(\mathbf{z}_{\mathrm{n}},\mathbf{z}_{\mathrm{f}})|_{\mathbf{z}^*}$ and is naturally interpretable as a Jacobian-driven update. If $M$ extracts a correlation between internal degrees of freedom and the observed change $\Delta \mathbf{z}$, then the update inherits the structure of linear response theory: one probes the system, measures the response, and uses that response to adapt parameters. This self-referential baseline subtraction also reduces bias from offsets or slow drifts in the operating point, making Mode~A especially natural for physical systems where free relaxation, such as energy minimization, is meaningful. Mode A does not require the target to appear directly in the perturbation step. Instead, the target enters implicitly through the choice of $M$. Below is the pseudo code for learning using Mode A. Perturbations can be implemented using either one-sided or symmetric nudging. 
In one-sided nudging, parameters are shifted as $\theta \rightarrow \theta + \delta\theta$, 
whereas symmetric nudging applies paired perturbations $\theta \pm \delta\theta$, 
enabling a central-difference estimate of the response that reduces bias and improves accuracy 
at the cost of additional measurements.
    
\begin{algorithm}[H]
\caption{Mode A: simple nudging}
\label{alg1}
\begin{algorithmic}[1]
\Require Dataset $\mathcal{D}$; feature map $F$; system response $G$; contrast measurement functional $M$;
initial parameters $\bm{\theta}_0$; learning rate $\eta$; perturbation rule for $\delta\bm{\theta}$; iterations $T$
\Statex \hspace{-\algorithmicindent}\rule{\linewidth}{0.4pt}
\State $\bm{\theta}\leftarrow \bm{\theta}_0$
\For{$i=0$ to $T-1$}
    \State $(\mathbf{x},\mathbf{z})\leftarrow \textsc{SampleBatch}(\mathcal{D})$
    \State $\mathbf{y} \leftarrow F(\mathbf{x})$
    \State \textbf{Free phase:} $\mathbf{z}_{\mathrm{f}} \leftarrow G(\mathbf{y},\bm{\theta})$
    \State \textbf{Nudge:} draw small $\delta\bm{\theta}$ and set $\bm{\theta}' \leftarrow \bm{\theta}+\delta\bm{\theta}$
    \State \textbf{Nudged phase:} $\mathbf{z}_{\mathrm{n}} \leftarrow G(\mathbf{y},\bm{\theta}')$
    \State \textbf{Contrast-to-update:} $\mathrm{corr} \leftarrow M(\mathbf{z}_{\mathrm{n}},\,\mathbf{z}_{\mathrm{f}})|_{\mathbf{z}^*}$
    \State \textbf{Parameter update:} $\bm{\theta} \leftarrow \bm{\theta} + \eta\,\mathrm{corr}$
\EndFor
\State \Return $\bm{\theta}$\end{algorithmic}
\end{algorithm}

\noindent 

Mode A provides a broader perturbative contrast framework that generalizes several existing physical learning algorithms. In Equilibrium Propagation (EP) \cite{scellier2017equilibrium}, learning arises from contrasts between free and weakly nudged equilibrium states in energy-based systems. Coupled Learning similarly extracts update signals from measurable differences between nearby physical responses with experiments implementing two copies of the system to determine differences\cite{stern2021supervised}, while Multi-mechanism learning eliminates the need for two copies by using two types of non-interfering responses, i.e. mechanical and electrical should they be decoupled~\cite{Anisetti2023}. Frequency Propagation considers different frequency domains~\cite{anisetti2024frequency}. Mode A unifies these approaches by treating learning as arising from measurable contrasts generated by controlled perturbations, independent of whether the underlying system is equilibrium-based, dynamical, classical, or quantum. In this view, the central object is not an explicit gradient or energy function, but the experimentally accessible response difference that acts as a physically measurable probe of the local response geometry of the system.

\subsection{Mode B: Target-referenced contrast: Anchored Learning}
Unlike Mode A, which compares nearby responses around the system’s free state, Mode B compares the system’s response to an externally specified target or imposed constraint. Assume we want to modify the weights $\bm{\theta}$ so that the response to a given $y$ is $z^{*}$. Linearizing the actual response about $\bm{\theta}$ gives 
\begin{align}
G(\mathbf{y},\bm{\theta}+\Delta\bm{\theta})
&\;\approx\;
G(\mathbf{y},\bm{\theta})
\;+\;
J_{\theta}\,\Delta\bm{\theta},\\
\mathbf{z}^*
&\;\approx\;
\mathbf{z}\;+\;
J_{\theta}\,\Delta\bm{\theta}.
\end{align}
Defining the instantaneous output error $\mathbf{e} \equiv \mathbf{z}^*-\mathbf{z}$, the linearized goal $\mathbf{e}\approx J_{\theta}\Delta\bm{\theta}$ is posed as a least-squares problem,
\begin{equation}
\Delta\bm{\theta}
\;=\;
\arg\min_{\Delta\bm{\theta}}
\left\| \mathbf{e} - J_{\theta}\Delta\bm{\theta}\right\|_2^2,
\label{eq:lsq}
\end{equation}
whose solution is the Moore--Penrose pseudo-inverse,
\begin{equation}
\Delta\bm{\theta}
\;\approx\;
J_{\theta}^{+}\,\mathbf{e},
\label{eq:pseudoinverse}
\end{equation}
or with Tikhonov regularization,
\begin{equation}
\Delta\bm{\theta}
\;\approx\;
\left(J_{\theta}^{\mathsf{T}}J_{\theta}+\lambda I\right)^{-1}
J_{\theta}^{\mathsf{T}}\,\mathbf{e}.
\label{eq:LM}
\end{equation}
Consequently, Mode~B does not merely probe local sensitivity (as in Mode~A), but instead performs a local, metric-aware correction step in parameter space, where the effective metric is given by $J_\theta^{\top} J_\theta$. Learning therefore follows the natural geometry induced by the system's input--output map rather than the Euclidean geometry of parameter space. Mode B performs target-referenced learning by translating output errors into parameter updates through the local response geometry of the system. Depending on the implementation, these updates may take the form of gradient-descent, Gauss–Newton, pseudoinverse, or other inverse-geometry corrections. However, in the implementations presented here, the Jacobian is measured through controlled perturbations and its pseudoinverse (or a regularized approximation) is computed externally. The key advantage of this metric-aware update is robustness: because $(J_\theta^\top J_\theta + \lambda I)^{-1}$ automatically normalizes the update direction according to the local response geometry, Mode~B remains stable across a wide range of learning rates and is insensitive to ill-conditioning of the input space in comparison with gradient descent, for example. Please see Appendix A for more details. Here is the pseudocode for Mode B.

\begin{algorithm}[H]
\caption{Mode B: direct inverse}
\label{alg2}
\begin{algorithmic}[1]
\Require Dataset $\mathcal{D}$; feature map $F$; system response $G$;
initial parameters $\bm{\theta}_0$; learning rate $\eta$; iterations $T$
\Statex \hspace{-\algorithmicindent}\rule{\linewidth}{0.4pt}
\State $\bm{\theta}\leftarrow \bm{\theta}_0$
\For{$i=0$ to $T-1$}
    \State $(\mathbf{x}, \mathbf{z}) \leftarrow \textsc{SampleBatch}(\mathcal{D})$
    \State $\mathbf{y} \leftarrow F(\mathbf{x})$
    \State \textbf{Reference output:} $\mathbf{z} \leftarrow G(\mathbf{y},\bm{\theta})$
    \State \textbf{Measure J:} vary $\bm{\theta}$ and evaluate $J_{\bm{\theta}}$ by measuring system response $\mathbf{z}$
    \State \textbf{Target mismatch:} $\mathbf{e} \leftarrow \mathbf{z}^* - \mathbf{z}$
    \State \textbf{Parameter update:} $\bm{\theta} \leftarrow \bm{\theta} + \eta\,J_{\bm{\theta}}^\dagger\,\mathbf{e}$
    \Statex \hspace{\algorithmicindent} \textit{(e.g.\ pseudoinverse or other invertible forms)}
\EndFor
\State \Return $\bm{\theta}$
\end{algorithmic}
\end{algorithm}

While Algorithm 2 presents Mode B in terms of an explicit local inverse, an alternative implementation can be constructed using perturbative contrasts and a measurement functional. In this approach, parameter updates are inferred from correlations between nudged responses and target outputs, allowing the effective inverse mapping to emerge from physically measurable quantities without requiring explicit evaluation of the Jacobian or its pseudoinverse.

\subsection{Two Learning Geometries: Sensitivity Probing vs. Local Inversion} Mode~A is most directly physical when the free response $\mathbf{z}_{\mathrm{f}}$ is a well-defined operating point (or steady state) and nudging corresponds to a small, controlled perturbation: the learning signal is then a bona fide linear response
measurement, Eq.~\eqref{eq:deltaG}. Mode~B becomes comparably physical when the target $\mathbf{z}^*$ is not a symbolic label but is embodied as a \emph{constraint}---a boundary condition, applied force pattern, clamped voltage, imposed concentration field, or geometric confinement. In that regime, the ``target'' is simply a physical condition the system is required to satisfy, and learning corresponds to remodeling internal parameters so that the constrained configuration becomes achievable (or stable) with progressively weaker external enforcement. A useful practical hybrid is therefore: employ Mode~B to \emph{imprint}
or \emph{instruct} desired behaviors under constraints, then use Mode~A to \emph{consolidate} them by aligning the free dynamics with the nudged behavior via contrastive, Jacobian-driven refinement. Regarding the measurement functional $M$, it abstracts the physically accessible measurement used to form a learning signal from pairs of system responses. In Mode~A, $M(\mathbf{z}_{\mathrm{f}},\mathbf{z}_{\mathrm{n}})$ extracts information proportional to the phase contrast
\begin{equation}
\Delta G \;\equiv\; \Delta\mathbf{z}=\mathbf{z}_{\mathrm{n}} - \mathbf{z}_{\mathrm{f}},
\end{equation}    
which, in the weak-perturbation limit, approximates the Jacobian action
$J_{\theta}\,\delta\bm{\theta}$ (Eq.~\ref{eq:deltaG}). In Mode~B, $M(\mathbf{z}_{\mathrm{n}},\mathbf{z}^*)$ encodes a local discrepancy between the nudged response (obtained via random perturbation or a physics-guided mechanism) and the target, which can be interpreted as an error signal driving a local inverse problem (Eq.~\ref{eq:lsq}).

While Mode B superficially resembles pseudoinverse-based learning rules used in machine learning, the underlying perspective is different. Conventional inverse methods begin with an explicit computational model and solve an inverse problem within that model. In contrast, Mode B infers the local learning geometry from physically measurable perturbations and response contrasts. In the proof-of-principle implementations presented here, this geometry is represented through an experimentally estimated Jacobian and an externally computed pseudoinverse update. As a result, Mode B can be interpreted as a physically realized pseudoinverse update, where the geometry of the parameter space is encoded in the system’s response to perturbations and read out through contrasts, rather than computed analytically, thereby realizing numerical optimization through physical response measurements. This distinction is crucial: the pseudoinverse update is an abstract linear-algebraic procedure, whereas Mode B implements metric-aware learning directly through physically measurable contrasts. Finally, Mode B is conceptually related to target propagation~\cite{lee2015difference}; however, target propagation constructs desired hidden-state targets using approximate inverse mappings, whereas Mode B directly solves a local inverse problem in parameter space using the measured response geometry of the physical system.

\section{Physical Realizations of PCPL in Mechanical and Photonic Systems}
We now present two concrete realizations of PCPL: a classical mechanical network and a continuous-variable photonic circuit. These examples are not intended as domain-specific optimizations, but as conceptual demonstrations of how the same perturbation–contrast principle gives rise to learning across different physical substrates. In particular, we emphasize how both systems naturally implement Mode B (target-referenced) learning through pseudoinverse or Gauss–Newton–like geometry, while remaining compatible with Mode A–style linear-response probing. See Appendix B for more details on PCPL. 

\subsection{Spring Networks as Contrastive Learners}

\begin{figure}[h]
    \centering
    \includegraphics[height=5cm]{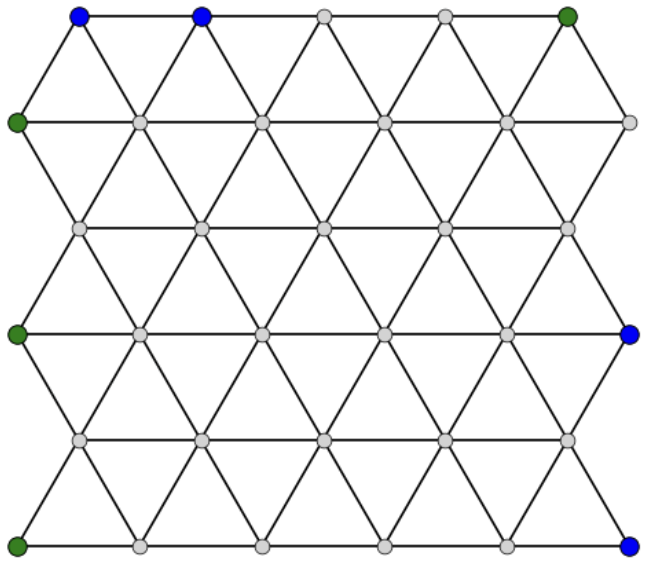}  
    \caption{\textit{Spring Network Schematic.} The green nodes are fixed and the 4 blue nodes denote where the input forces containing the Iris dataset information are applied.}
    \label{fig:triangular_lattice}
\end{figure}

\noindent{\bf The Network.} Consider a spring network with modifiable spring constants and equilibrium spring lengths. An illustrative example is shown in Fig.  \ref{fig:triangular_lattice}. Here, the blue nodes are the input ports, where data-dependent forces are applied, while the green nodes are fixed anchors that impose the boundary conditions. The rest of the network relaxes in response to these forces, producing a displacement field that serves as the physical response. The learning objective is to adjust mechanical parameters so that the displacement field describes a desired output given an input. The network will be trained to perform classification. We begin with the energy functional: 
\begin{align}
E &= \frac{1}{2}\sum_{ij}k_{ij}(l_{ij}-l_{ij}^{(0)})^2\notag\\&=\frac{1}{2}\sum_{ij}A_{ij} l_{ij}^2+\sum_{ij}B_{ij}l_{ij}+\frac{1}{2}\sum_{ij}C_{ij}\end{align}
where $A_{ij}=k_{ij},B_{ij}=-k_{ij}l_{ij}^{(0)},C_{ij}=k_{ij}l_{ij}^{(0)2}$. 

Given this energy functional, two training issues arise. First, to train rest length $l_{ij}^{(0)}$ and stiffness $k_{ij}$, we need to update $A,B$. Since $k_{ij}$ contributes to both $A$ and $B$, this creates an \emph{identifiability problem}: multiple combinations of $k_{ij}$ and $l_{ij}^{(0)}$ can produce the same energy, making the optimization ill-conditioned and prone to overfitting.

The second issue concerns gradient evaluation. In a mechanical network, the equilibrium node positions depend implicitly on all spring parameters through force balance,
\[
\nabla_{\mathbf u} E(\mathbf u,\mathbf k)=0 .
\]
Thus, although a partial derivative such as
\[
\frac{\partial E}{\partial k_{ij}}
=
\frac{1}{2}\left(l_{ij}-l^{(0)}_{ij}\right)^2
\]
appears local, changing \(k_{ij}\) shifts the equilibrium configuration \(\mathbf u^\ast(\mathbf k)\). More generally, for an objective \(\mathcal L(\mathbf u^\ast(\mathbf k),\mathbf k)\),
\[
\frac{d\mathcal L}{dk_{ij}}
=
\frac{\partial \mathcal L}{\partial k_{ij}}
+
\frac{\partial \mathcal L}{\partial \mathbf u}
\cdot
\frac{\partial \mathbf u^\ast}{\partial k_{ij}},
\]
so exact gradient computation requires the global sensitivity of the relaxed configuration to parameter changes. Linearizing the force-balance condition gives
\[
\frac{\partial \mathbf u^\ast}{\partial k_{ij}}
=
-
\left[
\frac{\partial^2 E}{\partial \mathbf u^2}
\right]^{-1}
\frac{\partial}{\partial k_{ij}}
\left(
\nabla_{\mathbf u}E
\right),
\]
which involves the inverse stiffness/Hessian matrix. To remain consistent with the PCPL philosophy and avoid this global adjoint calculation, we instead estimate parameter sensitivities from measured perturbative responses and construct updates from the resulting Jacobian. For the proof-of-principle example considered here, we restrict attention to a small linear spring network, where this Jacobian can be computed efficiently from the output displacements.

Given the first identifiability issue, we introduce coordinate vectors $\mathbf{u}_i$: setting the rest lengths as fixed to $l_{ij}^{(0)}(\mathbf{u}_i,\mathbf{u}_j)=1$ and add a gravitational force term. Thus, the energy equation becomes:
\begin{align}
E &= \frac{1}{2}\sum_{ij}k_{ij}(l_{ij}(u_i,u_j)-1)^2+g\sum_k m_k h_k(u_k)\notag\\
&=\frac{1}{2}\sum_{ij}k_{ij}(l_{ij}(u_i,u_j)^2-2l_{ij}(u_i,u_j)+1)\notag\\
&+g\sum_k m_k h_k(u_k)
\end{align}
where $h_k(u_k)$ represents the vertical height of node $k$. This approach enables us to separate the quadratic and linear terms, allowing us to train both spring constants $k_{ij}$ and masses $m_k$ simultaneously using the Jacobian pseudo-inverse method. To address the complexity of gradient computation, we restrict to a small linear spring network (Fig.~\ref{fig:triangular_lattice}), in which parameter updates are computed via the Jacobian of the output displacements with respect to the spring constants.

\noindent{\bf Iris dataset features as force inputs.}
We now demonstrate how a mechanical system equipped with a quadratic readout can implement target-referenced contrastive learning using a pseudoinverse update rule. While the classification task itself is standard, the purpose of this example is not to optimize accuracy but to show how a physical system can internalize a nonlinear decision boundary through Mode B geometry. We use a $6\times6$ triangular lattice as a linear spring network. The Iris features are encoded as dimensionless input vectors $\tilde{\mathbf{f}}\in\mathbb{R}^4$, which act as abstract forcing patterns rather than physical loads. Instead of interpreting these inputs as displacements or forces with physical units, we treat them as coordinates in a normalized input space and define a dimensionless objective function that couples the network’s mechanical response to a learned quadratic form. Specifically, the spring network produces a physical elastic energy $E_{\text{spring}}$, which we nondimensionalize by a reference energy scale $E_0$,
\begin{equation}
\tilde{E}_{\text{spring}} = \frac{E_{\text{spring}}}{E_0}.
\end{equation}
We then define a dimensionless quadratic functional
\begin{equation}
F(\tilde{\mathbf{f}}) = \tilde{\mathbf{f}}^\top W \tilde{\mathbf{f}} + \mathbf{b}^\top \tilde{\mathbf{f}} + c,
\end{equation}
where $W$, $\mathbf{b}$, and $c$ are dimensionless parameters learned during training. The total dimensionless readout is
\begin{equation}
\tilde{F}_{\text{total}} = F(\tilde{\mathbf{f}}) + \alpha \tilde{E}_{\text{spring}},
\end{equation}
where $\alpha$ is a dimensionless coupling coefficient that controls the relative contribution of the mechanical response. The two-dimensional feature space is $(\tilde{F}_{\text{total}}, \tilde{d}_{\max})$, where $\tilde{d}_{\max}$ is the normalized maximum displacement, is used for classification.

\noindent{\bf Training using PCPL Mode B.} Training is performed using a pseudoinverse learning rule. 
In particular, the Jacobian takes the form
\begin{equation}
J = \big[\mathrm{vec}(\tilde{\mathbf{f}}\tilde{\mathbf{f}}^\top),\ \tilde{\mathbf{f}},\ 1\big].
\end{equation}
To then determine the update $\Delta \bm{\theta}$, we use Tikhonov regularization in Equation \ref{eq:LM}. Finally, the parameters are updated using a learning rate $\eta$ as
\begin{equation}
\boldsymbol{\theta}_{\text{new}} = \boldsymbol{\theta} + \eta \, \Delta \boldsymbol{\theta},
\end{equation}
which is used to update the parameters $W$, $\mathbf{b}$, and $c$. In each epoch, the spring network is solved using the current input vector, producing $\tilde{E}_{\text{spring}}$, while the quadratic functional is evaluated using the same input. 

\noindent{\bf Classification thresholding.} We evaluated three decision strategies for classification, which differ in how the learned mechanical responses are mapped to discrete class labels:

\begin{itemize}
\item \textbf{Case 1: Scalar level matching in $F_{\text{total}}$.}  
Each sample is assigned to the class whose target value $t_c$ is closest to the observed scalar readout $F_{\text{total}}$, i.e.,
\[
\arg\min_c \left| F_{\text{total}} - t_c \right|.
\]
Here, the $t_c$ are class-specific reference values learned during training. Using this nearest-scalar readout rule, the system achieved $90.67\%$ test accuracy after 500 epochs with learning rate $\eta = 0.001$.

\item \textbf{Case 2: Adaptive boundary classification in scalar--displacement space.}  
 To map mechanical responses to discrete behavioral classes, we also employ an adaptive two-threshold classifier derived directly from the distribution of training data. For each input force pattern, the model computes a combined scalar response
\(
E_{\mathrm{tot}} = E_{\mathrm{FEM}} + \mathbf{f}^\top W \mathbf{f} + \mathbf{b}\cdot\mathbf{f} + c,
\)
where \(E_{\mathrm{FEM}}\) is the physical strain energy from the lattice simulation and the quadratic form represents the learned energetic contribution. In parallel, we extract the maximum nodal displacement \(D_{\max}\) from the mechanical solution. Class boundaries are then determined adaptively from training-set extrema rather than fixed parameters. Specifically, an energy threshold
\[
E^\ast = \tfrac{1}{2}\big(\min E_0 + \max(E_1, E_2)\big)
\]
separates class 0 from the union of classes 1 and 2. Within the low-energy regime \(E_{\mathrm{tot}} \le E^\ast\), a second threshold
\[
D^\ast = \tfrac{1}{2}\big(\max D_1 + \min D_2\big)
\]
distinguishes class 1 from class 2. The resulting decision rule is therefore hierarchical: energy first partitions responses into “high-energy” versus “low-energy” behaviors, after which displacement resolves the low-energy cases into distinct mechanical states. Because both thresholds are recomputed from the current training distribution, the decision boundary adapts automatically to shifts in system stiffness, geometry, or learned parameters.

\item \textbf{Case 3: Mechanically grounded scalar classification.}  In this strategy, the learned quadratic readout is removed, and classification is based solely on a scalar summary of the spring network’s mechanical response. For each input force pattern, the network is solved to mechanical equilibrium, producing a displacement field $u$. From this state, we compute a dimensionless scalar measure of deformation,
\[
S_{\mathrm{mech}} = \tfrac{1}{2}\, u^\top K u,
\]
which reflects the global elastic response of the lattice rather than a direct algebraic function of the inputs. Because the equilibrium displacements arise from a coupled force-balance problem, the mapping from input features to $S_{\mathrm{mech}}$ is geometrically nonlinear, even though the energy expression itself is quadratic in $u$. Class boundaries are determined adaptively from gaps in the training distributions of $S_{\mathrm{mech}}$. Specifically, two scalar thresholds are defined as
\[
b_1 = \tfrac{1}{2}\big(\max S_0 + \min S_1\big), \qquad
b_2 = \tfrac{1}{2}\big(\max S_1 + \min S_2\big),
\]
where $S_i$ denotes the set of mechanical scalar responses for class $i$. The resulting decision rule is therefore a two-level threshold classifier operating in a one-dimensional, mechanically generated feature space. Because the thresholds are recomputed from the current training distribution, the classifier adapts automatically to changes in stiffness, geometry, and learned parameters while remaining grounded in physically measurable quantities.
\end{itemize}

\begin{figure}[h]
    \centering
    \begin{tabular}{c}
    \includegraphics[height=6.5cm]{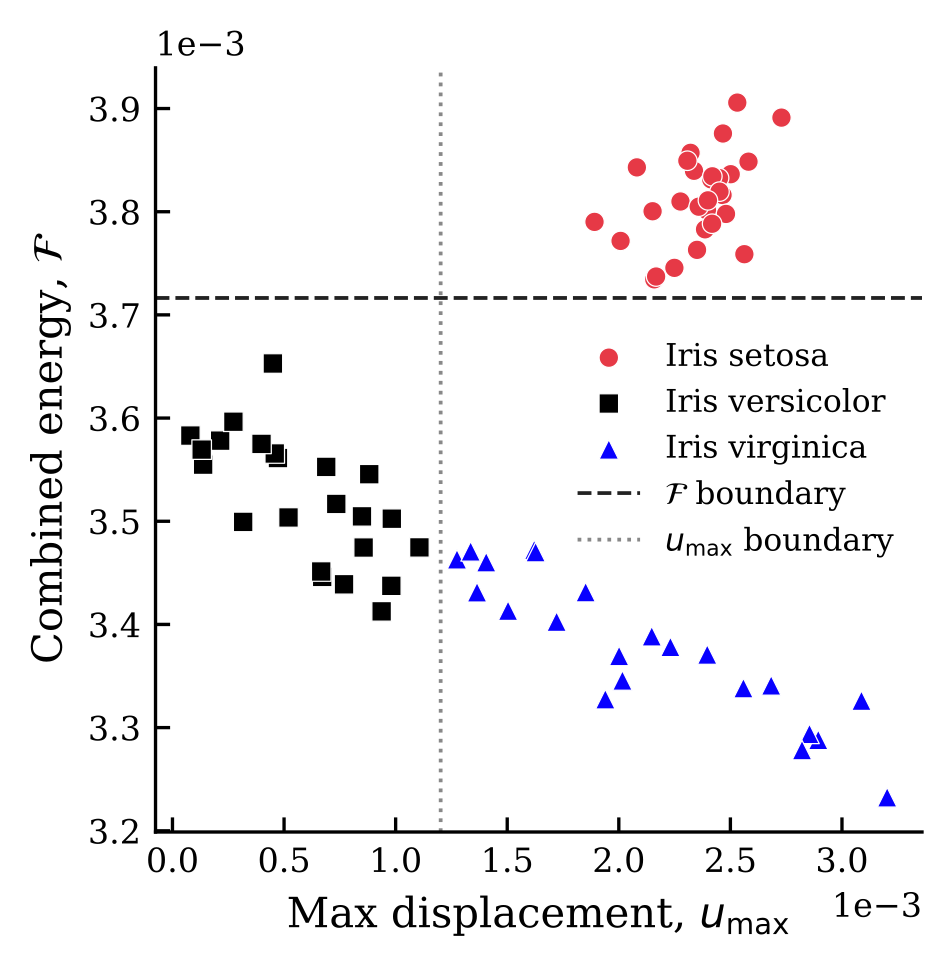}  \\
    \includegraphics[height=6.5cm]{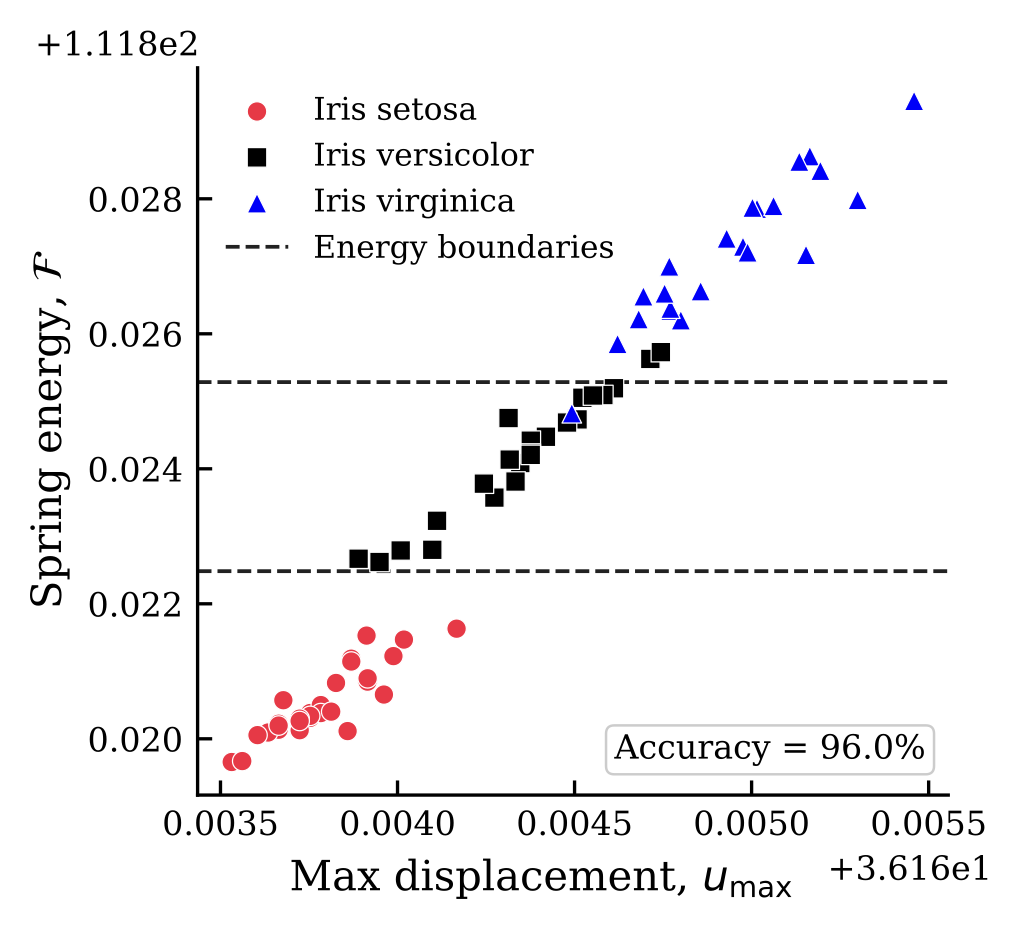} \\
    \end{tabular}
    \caption{\textit{Decision boundaries of case 2 and case 3 for a classical spring network in a gravitational field:} Left - Hybrid system classification using adaptive thresholds in $F_{\text{total}}$ vs. displacement space achieves 100\% test accuracy. Right - Pure mechanical system classification using $E_{\text{total}}$ (linear strain energy) with gap-based energy thresholds achieves 96\% accuracy, where both stiffness matrix $K$ and masses $\mathbf{b}$ are trained via Jacobian pseudoinverse with parameter perturbation.}
    \label{fig:figureEF}
\end{figure}

\noindent{\bf Performance.} Figure \ref{fig:figureEF} shows the decision boundary of Case 2 and Case 3, which both yielded higher accuracy than Case 1. The difference in classification accuracy across Cases 1–3 reflects the geometric expressivity of the feature spaces rather than merely the number of thresholds. Case 1 relies on a single scalar readout with nearest-level matching, which restricts the decision rule to affine separations and yields lower accuracy. In contrast, Cases 2 and 3 both employ two-threshold decision rules, but operate in nonlinear feature spaces generated by the physical system. Case 2 uses a two-dimensional feature space $(E_{\mathrm{tot}}, D_{\max})$, enabling separations that exploit independent mechanical response modes and thereby achieving the highest accuracy. Case 3, while also nonlinear, compresses all class information into a single mechanically generated scalar, limiting separability when class clusters overlap along that axis and resulting in slightly reduced performance relative to Case 2.

Interestingly, using a spring network physical learning system with linear classification, prior work reports an average classification accuracy of $96\%$ on the Iris dataset using a $70/30$ train--test split~\cite{Li2024mechanical}. Under slightly more challenging splits, such as $50/50$, this performance may be expected to decrease further due to both the reduced amount of training data and the restricted representational capacity of linear models, which can only implement affine input--output mappings of the form $\mathbf{A}\mathbf{y} + \mathbf{b} = \mathbf{z}$. Such mappings are unable to represent curved or topologically nontrivial decision boundaries. This limitation is reflected in our \textbf{Case~1}, which relies on a scalar-level matching rule in a one-dimensional readout space and therefore inherits the expressivity constraints of linear classifiers. 

As expected, geometrically nonlinear decision rules can achieve substantially higher performance on this dataset. Indeed, a wide range of nonlinear models report test accuracies approaching $\sim 98\%$ \cite{Nugroho2020}, which appears to represent a practical generalization ceiling imposed by intrinsic class overlap. Our \textbf{Case~2}, which constructs adaptive decision boundaries in the two-dimensional $(F_{\text{total}}, D)$ feature space, falls into this category: by allowing curved and data-adaptive separations, it achieves $100\%$ test accuracy for particular train--test partitions. Importantly, such perfect separation should be interpreted as a finite-sample geometric effect rather than evidence that the intrinsic limit has been exceeded. For specific splits, the learned boundaries may align favorably with the data geometry, but this behavior is not expected to persist under repeated randomized splits or cross-validation. 

Finally, \textbf{Case~3} introduces an additional layer of nonlinearity through the mechanically-mediated mapping between inputs and scalar readouts, allowing the classifier to exploit structure induced by the spring network itself. While still subject to the same intrinsic data overlap constraints, this mechanically grounded nonlinearity provides a distinct route to enhanced expressivity that differs fundamentally from purely algebraic nonlinear classifiers. Together, these results highlight the tradeoff between computational simplicity and geometric expressivity, and clarify the distinction between split-specific performance and intrinsic generalization limits.

\subsection{Continuous Variable Photonic Circuits as Contrastive Learners}

We now show that the same PCPL framework applies to continuous-variable (CV) photonic systems. For convenience, we employ the standard Gaussian-state formalism of continuous-variable optics and represent optical modes using bra-ket notation. Because the examples considered here remain entirely within the Gaussian sector, they also admit an equivalent description in terms of classical optical field amplitudes. The learning logic, however, remains unchanged: perturb, contrast, and update. Let us build continuous variable photonic circuits out of the following components: displacement gates $D(\alpha)$,  phase shift, a.k.a.\ phase-space rotation, gates $R(\phi_i)$, 
beam splitter gates, $\text{BS}(\theta, \phi)$ to create two-mode couplings, and squeezing gates. The circuit will contain modifiable parameters so that it can learn to classify objects, for example.  Using our learning notation, $F$ prepares photonic states ($F: \mathbf{x} \mapsto \ket{\psi(\mathbf{x})}$), $G$ applies parameterized operations: $G: \ket{\psi(\mathbf{x})}, \bm{\theta} \mapsto \ket{\psi(\mathbf{x}; \bm{\theta})}$. 

As the parameter contrast in Mode A~\cite{Scellier2024} is given by $\Delta \bm{\theta} = M(\psi_\mathrm{free},\psi_\mathrm{nudged})$ depending on the readout, the update becomes one of the following, 
\begin{align}
\Delta \theta_\text{phase}^i &= \alpha_i \cdot (\langle \hat{X}_i \rangle_\text{nudged} - \langle \hat{X}_i \rangle_\text{free}) ,\\
\Delta \theta_\text{sq}^i &= (\text{Var}(\hat{X}_i)_\text{nudged} - \text{Var}(\hat{X}_i)_\text{free}) ,\\
\Delta \theta_\text{bs}^{ij} &= (\text{Cov}(\hat{X}_i, \hat{X}_j)_\text{nudged} - \text{Cov}(\hat{X}_i, \hat{X}_j)_\text{free}).
\end{align}
Similarly, a Mode B (target-referenced) contrast can be written generally as $\Delta \bm{\theta}
=
M(\psi_\mathrm{target},\psi_\mathrm{nudged})$ with the following possible readouts
\begin{align}
\Delta \theta_{\mathrm{phase}}^{i}
&= \alpha_i\Big(\langle \hat{X}_i\rangle_{\mathrm{nudged}}-\langle \hat{X}_i\rangle_{\mathrm{target}}\Big),\\
\Delta \theta_{\mathrm{sq}}^{i}
&= \Big(\mathrm{Var}(\hat{X}_i)_{\mathrm{nudged}}-\mathrm{Var}(\hat{X}_i)_{\mathrm{target}}\Big),\\
\Delta \theta_{\mathrm{bs}}^{ij}
&= \Big(\mathrm{Cov}(\hat{X}_i,\hat{X}_j)_{\mathrm{nudged}}-\mathrm{Cov}(\hat{X}_i,\hat{X}_j)_{\mathrm{target}}\Big).
\end{align}

\subsubsection{Gaussian CV photonic (using linear optical gates) circuit}
This example demonstrates how Mode~B PCPL can be implemented in a CV photonic circuit to perform Iris classification. Classical input features are encoded into Gaussian optical states, and learning proceeds through measurement contrasts that drive pseudoinverse (Gauss–Newton–like) parameter updates. 
	\begin{figure*}[]
		\centering
		\begin{tikzpicture}[scale=0.6]
			\node[scale=0.7] {
				\begin{quantikz}[column sep=0.5cm, row sep=0.4cm,transparent]
					\lstick{$\vert 0\rangle_0$}
					& \gate{D(x_1)}
					& \gate{R(\psi_1)}
					& \gate[2]{U_{\mathrm{L1}}:\,BS_{01}(\theta_1,\phi_1)}
					& \qw
					& \gate[3,label style={yshift=0.4cm}]{U_{\mathrm{L2}}:\,BS_{02}(\theta_3,\phi_3)}
					& \qw
					& \gate[4]{U_f:\,BS_{03}\!\left(\tfrac{\pi}{4},0\right)}
					& \gate[2]{U_f:\,BS_{01}\!\left(\tfrac{\pi}{4},0\right)}
					& \qw
					& \meterD{x_0} \\
					\lstick{$\vert 0\rangle_1$}
					& \gate{D(x_2)}
					& \gate{R(\psi_2)}
					& \qw
					& \gate[3,label style={yshift=0.5cm}]{U_{\mathrm{L2}}:\,BS_{13}(\theta_4,\phi_4)}
					& \linethrough\qw
					& \qw
					& \linethrough\qw
					& \qw
					& \qw \\
					\lstick{$\vert 0\rangle_2$}
					& \gate{D(x_3)}
					& \gate{R(\psi_3)}
					& \gate[2]{U_{\mathrm{L1}}:\,BS_{23}(\theta_2,\phi_2)}
					& \linethrough\qw
					& \qw
					& \qw
					& \linethrough\qw
					& \qw
					& \qw \\
					\lstick{$\vert 0\rangle_3$}
					& \gate{D(x_4)}
					& \gate{R(\psi_4)}
					& \qw
					& \qw
					& \qw
					& \qw
					& \qw
					& \qw
					& \qw
				\end{quantikz}
			};
		\end{tikzpicture}
		\caption{\textit{Four-mode Gaussian photonic circuit for linear classification.}
			Each mode is initialized in vacuum $\vert 0\rangle$, displaced by $D(x_i)$,
			and rotated by $R(\phi_i)$, giving $U_R = \mathrm{diag}(e^{i\phi_1},\ldots,e^{i\phi_4})$. This can be implemented by phase modulating lasers of tunable power.
			The first beam splitter layer
			$U_{\mathrm{L1}} = BS_{01}(\theta_1,\phi_1) \oplus BS_{23}(\theta_2,\phi_2)$
			and second layer
			$U_{\mathrm{L2}} = BS_{02}(\theta_3,\phi_3) \oplus BS_{13}(\theta_4,\phi_4)$
			mix the modes using 8 trainable parameters.
			The fixed concentration layer
			$U_f = BS_{03}(\pi/4,0)\cdot BS_{01}(\pi/4,0)$
			concentrates information into mode~0.
			$x$-quadrature mean measurements yields
			$\langle\hat{x}_0\rangle = \sum_i W_i x_i$,
			where $W_i = \sqrt{2}\,\mathrm{Re}[(U_{\mathrm{BS}})_{0i}\,e^{i\phi_i}]$
			and $U_{\mathrm{BS}} = U_f\,U_{\mathrm{L2}}\,U_{\mathrm{L1}}$.}
\label{fig:cv-photonic-circuit}
	\end{figure*}
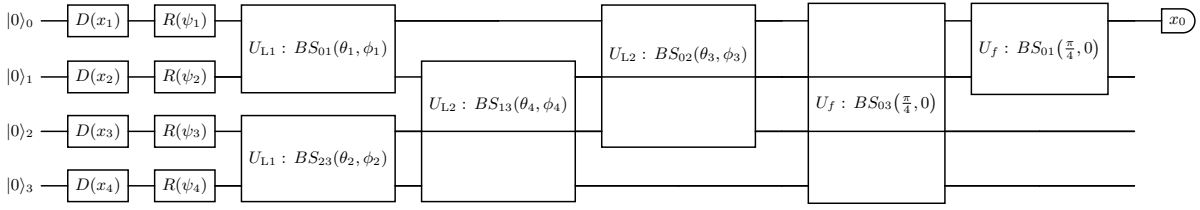

\noindent{\bf Circuit architecture and component operations.} The input feature vector $\mathbf{x} \in \mathbb{R}^4$ is encoded into a four-mode photonic system using displacement gates $D(x_i)$ applied to the vacuum state. These displacements shift the expectation values of the position quadratures, embedding classical data into photonic phase space. The displaced modes then pass through parameterized Gaussian operations consisting of rotation gates $R(\psi_i)$ and a network of beam splitter gates
$BS_{ij}(\theta_k,\phi_k)$ that mix quadratures and couple modes. Finally, a set of beam splitters concentrates the information into a single mode, and the expectation value of its position quadrature $\langle \hat{x}_0 \rangle$ is measured via $x$-quadrature mean measurements. See Fig. \ref{fig:cv-photonic-circuit}. Let us more specifically analyze how the sequence of Gaussian operations
transforms $\langle \hat{x}_0 \rangle$.

\begin{enumerate}
	
	\item \textbf{Initial State}: The vacuum state has
	$\langle \hat{x}_i \rangle = \langle \hat{p}_i \rangle = 0$, or equivalently
	$\alpha_i \equiv \langle \hat{a}_i \rangle = 0$, for all modes $i \in \{0,1,2,3\}$.
	
	\item \textbf{Displacement Operation}: The amplitude  displacement gate $D_i(x_i \in \mathbb{R}) = \exp(-i\sqrt2\,x_iP_i)$, applied to each mode $i$, shifts the initial amplitude $\alpha_i$:
	\begin{align}
		\alpha_i \rightarrow \alpha_i + x_i
	\end{align}
	so that after displacement, $\alpha_i$ becomes $0+x_i$.
	
	\item \textbf{Rotation Operation}: The rotation gate
	$R_i(\psi_i) = \exp(i\psi_i \hat{n}_i)$ acts on the complex amplitude as a phase shift:
	\begin{align}
		\alpha_i \rightarrow e^{i\psi_i}\,\alpha_i = e^{i\psi_i}\,x_i.
	\end{align}
	Collectively across all four modes, this defines the diagonal unitary
	\begin{align}
		U_R = \mathrm{diag}(e^{i\psi_0}, e^{i\psi_1}, e^{i\psi_2}, e^{i\psi_3}) \in U(4)
	\end{align}
	so that $\vec{\alpha} \rightarrow U_R\,\mathbf{x}$.
	The rotation angles $\{\psi_i\}_{i=0}^{3}$ therefore control the complex phase of each
	mode's contribution to the output.
		\item \textbf{Beam Splitter Network}: The beam splitter
	 $BS_{ij}(\theta_k, \phi_k)$ acts as a
	$U(2)$ rotation on the pair $(\alpha_i, \alpha_j)$:
	\begin{align}
		\begin{pmatrix}\alpha_i \\ \alpha_j\end{pmatrix}
		\;\to\;
		\begin{pmatrix}
			\cos\theta_k & -e^{i\phi_k}\sin\theta_k \\
			e^{-i\phi_k}\sin\theta_k & \cos\theta_k
		\end{pmatrix}
		\begin{pmatrix}\alpha_i \\ \alpha_j\end{pmatrix}.
	\end{align}
	Its extension to the full four-mode space is:
	\begin{align}
		\widetilde{BS}_{ij}(\theta_k,\phi_k) =
		\begin{pmatrix} BS_{ij}(\theta_k,\phi_k) & 0 \\ 0 & I_2
		\end{pmatrix}_{(i,j,k,l)} \in U(4),
	\end{align}
	where the subscript $(i,j,k,l)$ indicates the row/column ordering with
	$\{k,l\} = \{0,1,2,3\}\setminus\{i,j\}$ being the unchanged modes.
	The layer unitaries are then products of these extended operators:
	\begin{align}
		U_{\mathrm{L1}} &= \widetilde{BS}_{01}(\theta_1,\phi_1) \cdot
		\widetilde{BS}_{23}(\theta_2,\phi_2) \in U(4) \\[4pt]
		U_{\mathrm{L2}} &= \widetilde{BS}_{02}(\theta_3,\phi_3) \cdot
		\widetilde{BS}_{13}(\theta_4,\phi_4) \in U(4) \\[4pt]
		U_f             &= \widetilde{BS}_{03}\!\left(\tfrac{\pi}{4},0\right) \cdot
		\widetilde{BS}_{01}\!\left(\tfrac{\pi}{4},0\right) \in U(4)
	\end{align}
	
\end{enumerate}

The full circuit applies the following sequence of operations to the
initial state $\vec{\alpha} = \mathbf{x}$ (see Fig. \ref{fig:cv-photonic-circuit}):
\begin{align}
	\vec{\alpha}^{\,\mathrm{out}}
	= U_f\, U_{\mathrm{L2}}\, U_{\mathrm{L1}}\, U_R\, \mathbf{x}
	\equiv U_{\mathrm{BS}}\, U_R\, \mathbf{x},
\end{align}
where $U_{\mathrm{BS}} \equiv U_f\, U_{\mathrm{L2}}\, U_{\mathrm{L1}} \in U(4)$.
Reading off the zeroth component, with $\mathbf{e}_0 = (1,0,0,0)^\top$:
\begin{align}
	\alpha_0^{\mathrm{out}}
	= \bigl(\mathbf{e}_0^\top\, U_{\mathrm{BS}}\, U_R\bigr)\, \mathbf{x}
	= \sum_{i=0}^{3} \bigl(U_{\mathrm{BS}}\bigr)_{0i}\, e^{i\phi_i}\, x_i.
\end{align}
The $x$-quadrature measurement of mode~0 then yields:
\begin{align}
	\langle \hat{x}_0 \rangle
	&= \sqrt{2}\,\mathrm{Re}\!\left(\alpha_0^{\mathrm{out}}\right)
	= \sqrt{2}\sum_{i=0}^{3}
	\mathrm{Re}\!\left[\bigl(U_{\mathrm{BS}}\bigr)_{0i}\, e^{i\phi_i}\right] x_i\notag\\
	&\equiv \sum_{i=0}^{3} W_i\, x_i,
\end{align}
where the effective classifier weights are:
\begin{align}
	W_i = \sqrt{2}\,\mathrm{Re}\!\left[\bigl(U_{\mathrm{BS}}\bigr)_{0i}\, e^{i\phi_i}\right].
\end{align}
The 12 trainable parameters $\{\phi_i\}_{i=0}^{3} \cup \{\theta_k,\phi_k\}_{k=1}^{4}$
jointly determine the effective linear classifier weights $\{W_i\}$ through the
composite map $U_{\mathrm{BS}}\, U_R$. Consequently, the circuit output can be
viewed as an inner product:
\begin{align}
	\langle \hat{x}_0 \rangle
	= \mathbf{W} \cdot \mathbf{x}
	= \sum_{i=0}^{3} W_i\, x_i
\end{align}
where $\mathbf{W} = (W_0, W_1, W_2, W_3)^\top \in \mathbb{R}^4$ is the effective
weight vector and $\mathbf{x} \in \mathbb{R}^4$ is the input feature vector.
Despite operating on photonic states, the measured output is thus a
\emph{linear function} of the classical inputs, with the photonic circuit
parameterizing the weight vector $\mathbf{W}$ through the unitary degrees of freedom.

\noindent{\bf Nonlinearity via Jacobian pseudoinverse updates.} Although the instantaneous circuit output is linear in the inputs,
$\langle \hat{x}_0 \rangle = \mathbf{W}(\boldsymbol{\theta}) \cdot \mathbf{x}$,
the parameter update rule introduces an effective nonlinearity.
The Jacobian matrix $J \in \mathbb{R}^{N \times 12}$, stacked over a batch of
$N$ samples, has entries:
\begin{align}
	J_{ki} = \frac{\partial \langle \hat{x}_0 \rangle^{(k)}}{\partial \theta_i}
	= \frac{\partial \mathbf{W}(\boldsymbol{\theta})}{\partial \theta_i}
	\cdot \mathbf{x}^{(k)}
	\equiv \mathbf{g}_i(\boldsymbol{\theta}) \cdot \mathbf{x}^{(k)},
\end{align}
where $\mathbf{g}_i(\boldsymbol{\theta}) = \partial \mathbf{W}/\partial\theta_i$
is generally nonlinear in $\boldsymbol{\theta}$ through the trigonometric
dependence of $U_{\mathrm{BS}}$. Moreover, there are other sources of nonlinearity.  First, if the true input-output relationship is nonlinear, the stacked system $J\,\Delta\boldsymbol{\theta} = \mathbf{e}$ is overdetermined and inconsistent --- no single linear weight vector $\mathbf{W}$ can satisfy all $N$ equations simultaneously. The pseudoinverse solution then finds a least-squares compromise across the batch, whose effective behavior depends nonlinearly on the data distribution $\{(\mathbf{x}^{(k)}, \mathbf{e}^{(k)})\}$. Second, given that the update rule for Mode B contains $J^\top J$, the entries
	\begin{align}
		(J^\top J)_{ij}
		= \sum_{k=1}^{N}
		\bigl(\mathbf{g}_i \cdot \mathbf{x}^{(k)}\bigr)
		\bigl(\mathbf{g}_j \cdot \mathbf{x}^{(k)}\bigr)
	\end{align}
	are \emph{quadratic} in $\mathbf{x}^{(k)}$. The matrix inverse
	$(J^\top J + \lambda I)^{-1}$ then promotes this to a rational function
	of $\{\mathbf{x}^{(k)}\}$, analogous to a polynomial kernel
	$K(\mathbf{x},\mathbf{x}') = (\mathbf{x} \cdot \mathbf{x}')^2$.
Therefore, although $J_{ki}$ is linear in $\mathbf{x}^{(k)}$, the \emph{multiplication of two such linear terms} in $J^\top J$
that breaks linearity of the update rule, while the mini-batch overdetermination additionally encodes the nonlinear structure of the data distribution. In sum, the learned input-output map remains linear in the input features once training is complete. The nonlinearity enters the learning dynamics rather than the inference map.

\noindent{\bf Mini-batch implementation.} Training proceeds using mini-batches of input samples, each encoded into a separate preparation of the photonic circuit. For a batch of size $N$, the Jacobian becomes an $N \times q$ matrix, and the pseudoinverse update integrates response information across multiple inputs simultaneously. This improves numerical conditioning of the update and reduces the influence of measurement noise, while remaining compatible with sequential experimental execution in hardware implementations. 

In our implementation, a mini-batch of size 5 corresponds to evaluating five independent four-mode Gaussian circuits. Each input $\mathbf{x} \in \mathbb{R}^4$ is encoded via displacement gates across four modes. Denoting $z_{kj}^{(+)} = \langle\hat{x}_0\rangle_k(\theta_j + \varepsilon)$ and $z_{kj} = \langle\hat{x}_0\rangle_k(\theta_j)$ as the perturbed and unperturbed $x$-quadrature mean outputs for the $k$-th sample and $j$-th parameter, the Jacobian entries can be directly expressed using the bilinear form:
\begin{align}
J_{kj} = \frac{\partial \langle \hat{x}_0 \rangle_k}{\partial \theta_j} \simeq \frac{M(z_{kj}^{(+)}, z_{kj})}{(\theta_j+\varepsilon)-\theta_j}&=  \frac{z_{kj}^{(+)} - z_{kj}}{\varepsilon}\\\notag
&= \sum_{i=0}^3 P_{ij} \cdot x_{ki}
\end{align}

where $x_{ki}$ is the $i$-th feature of the $k$-th sample. This makes clear that $P$ emerges implicitly from the physical measurements without ever being explicitly constructed. For a batch of $N$ samples, we can write the full Jacobian matrix as: 
\begin{align}
J = X P
\end{align}
where $X \in \mathbb{R}^{N \times 4}$ is the input data matrix and $P \in \mathbb{R}^{4 \times q}$ contains the fixed coefficients $P_{ij}$.

Using this notation, the pseudoinverse update becomes:
\begin{align}
\Delta \boldsymbol{\theta} &= (J^T J + \lambda I)^{-1} J^T \mathbf{e} \\
&= ((X P)^T (X P) + \lambda I)^{-1} (X P)^T \mathbf{e} \\
&= (P^T X^T X P + \lambda I)^{-1} P^T X^T \mathbf{e}
\end{align}
where $\mathbf{e} = \mathbf{z}^* - \mathbf{z}$ is the error vector between the target and $x$-quadrature mean outputs, as defined previously.

During training, the parameter vector $\boldsymbol{\theta}$ traces a trajectory through the parameter space $\mathbb{R}^q$. This trajectory is guided by:
\begin{align}
\boldsymbol{\theta}_{t+1} = \boldsymbol{\theta}_t + \eta \Delta \boldsymbol{\theta}_t.
\end{align}
The learning rate $\eta$ controls the step size along this trajectory. 

\noindent{\bf Classification from continuous measurements.} Because the
circuit produces a continuous scalar output $\langle \hat{x}_0 \rangle$,
classification is performed by mapping this value into discrete labels
using two adaptive thresholds $t_1 < t_2$. At each training step, the
thresholds are updated as the midpoints between the mean outputs of
adjacent classes,
\[
t_k \leftarrow (1 - \gamma)\,t_k + \gamma\,\frac{\mu_k + \mu_{k+1}}{2},
\quad k = 1, 2,
\]
where $\mu_k$ denotes the mean measured output for class $k$ within the
current mini-batch and $\gamma > 0$ is a damping factor. The decision
rule is then
\[
\text{class} =
\begin{cases}
0, & \langle \hat{x}_0 \rangle < t_1,\\
1, & t_1 \le \langle \hat{x}_0 \rangle < t_2,\\
2, & \langle \hat{x}_0 \rangle \ge t_2.
\end{cases}
\]
This adaptive scheme ensures the thresholds track shifts in the
circuit's learned response while remaining grounded in physically
measurable observables.

\begin{figure}[h]
    \centering
    \begin{tabular}{l}
                \includegraphics[height=4.5cm]{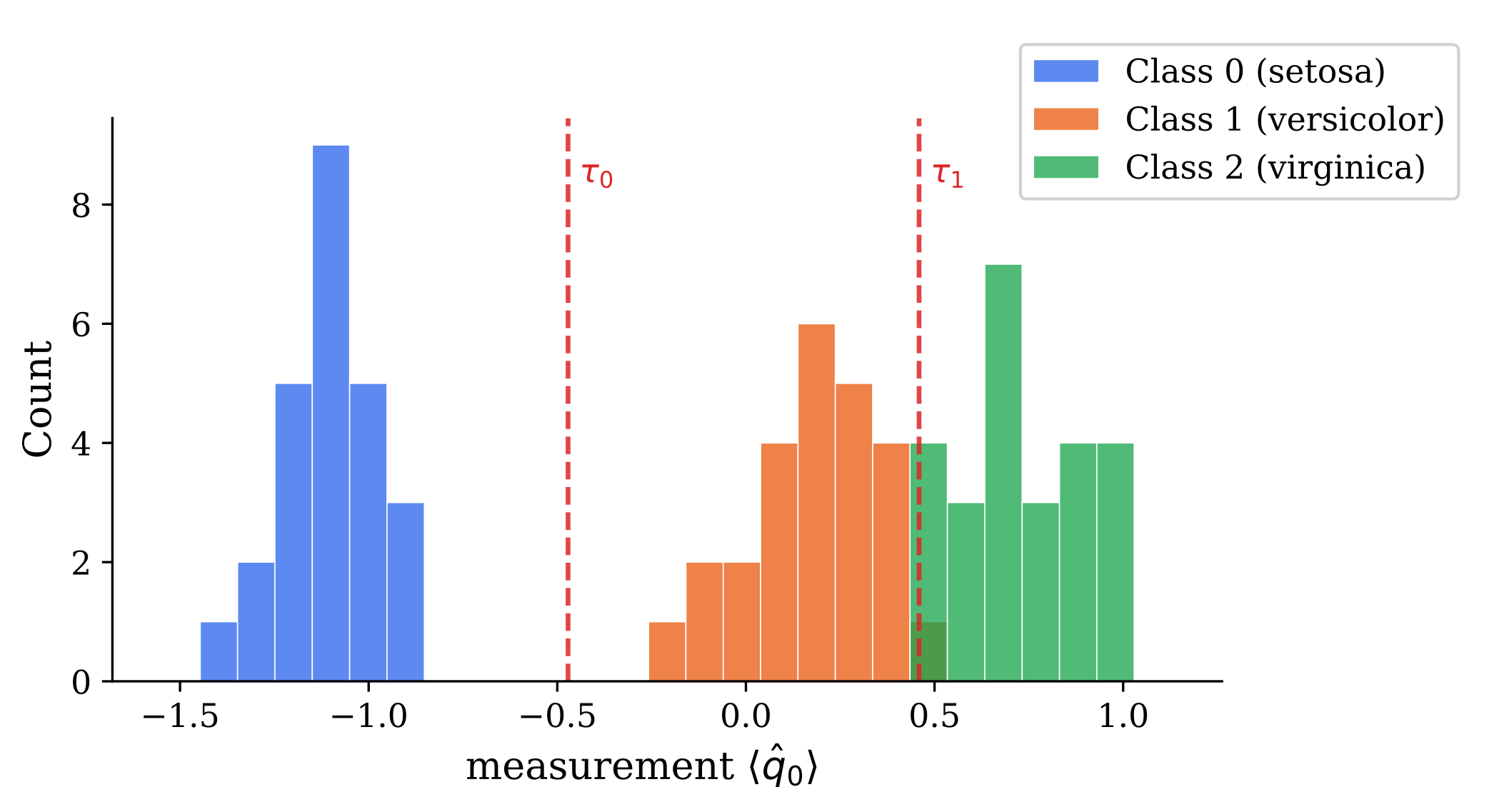}  \\  
         (a) Best Classification ($100.0\%$) \\\includegraphics[height=4.5cm]{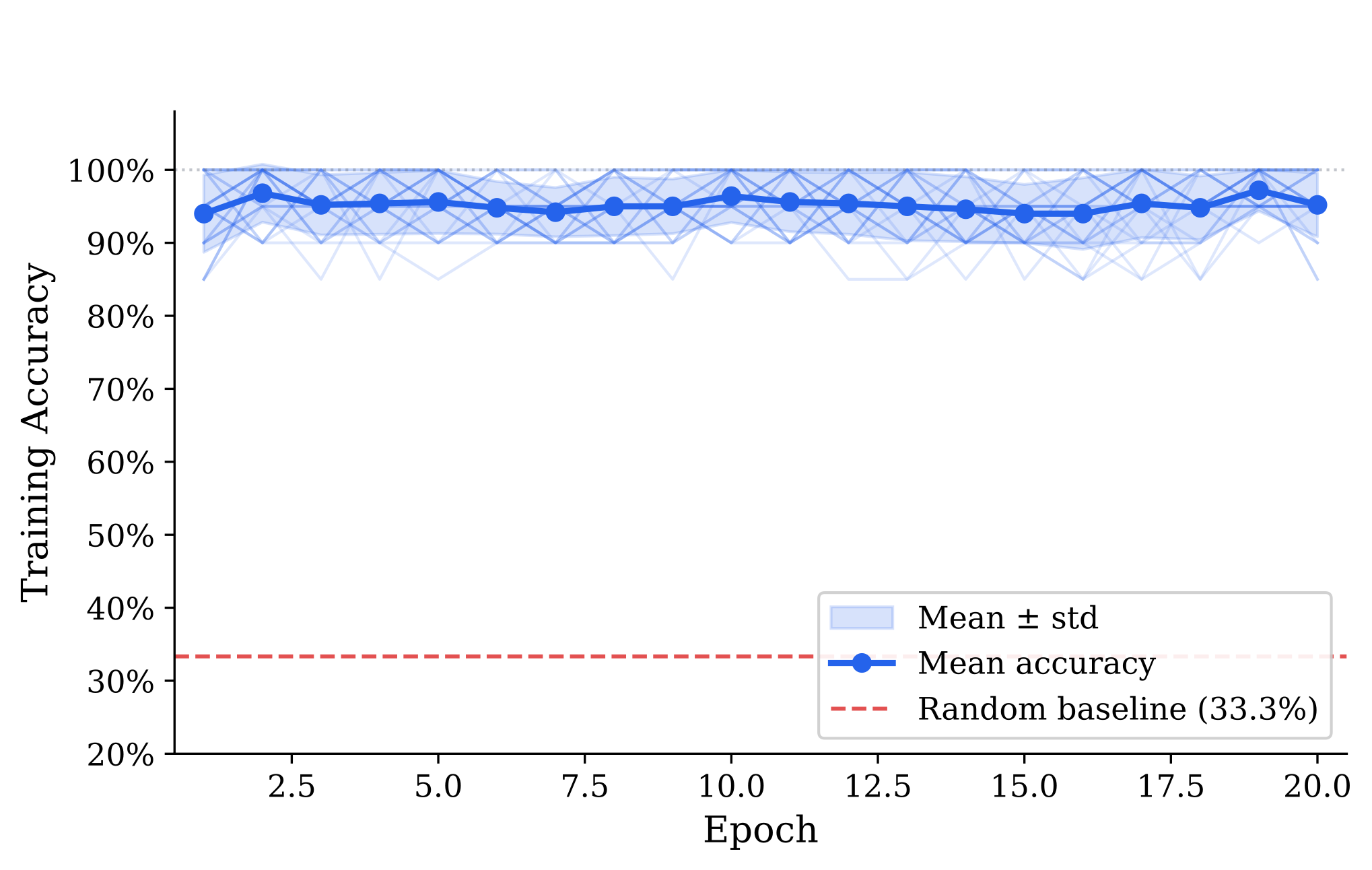} \\ (b) Batch size $N=5$, $\eta=0.1$
    \end{tabular}
    \caption{{\it Classification accuracy via mini-batch pseudoinverse nudging.} (a) distribution of photonic measurement outcomes by class showing clear separation between Class 0 (blue, centered at -1), Class 1 (orange, centered at 0), and Class 2 (green), achieving 100.0\% classification accuracy. (b) Training convergence using mini-batch pseudoinverse nudging with batch size $N=5$ and learning rate $\eta=0.1$, reaching nearly perfect accuracy within 10 epochs.}
    \label{fig:cv-photonic-performance}
\end{figure}

\noindent{\bf Performance.} With a 50\%/50\% train/test split over 25 independent trials, the test accuracy ranges from $96.0\%$ to $100.0\%$, with a mean of $97.7\% \pm 0.9\%$ (20 epochs). See Fig. \ref{fig:cv-photonic-performance}. Notably, this performance
is achieved using only Gaussian operations and $x$-quadrature
measurements, without invoking intrinsic photonic nonlinearities. The
classifier's expressivity arises from the geometry of the circuit's
input--output map and the conditioning of its Jacobian under parameter
perturbations. The pseudoinverse update succeeds because its physically realizable response
manifold supports well-conditioned Gauss--Newton--like corrections. This
demonstrates that PCPL learning performance depends primarily on how a
physical system structures parameter sensitivities, rather than on the
presence of specific computational primitives. More generally, PCPL is compatible with nonlinear circuitry, including squeezing, and, in principle, other quantum measurement schemes, including photon-number detection, so long as measurable contrasts between nearby responses can be used to generate update signals. 

\noindent{\bf Changing circuit architecture.} To understand what aspects of the circuit enable effective Mode B learning, we systematically altered the circuit architecture and examined how these changes affected learning performance. First, removing the final mode-concentration stage did not significantly affect accuracy. This indicates that compressing information into a single mode is not essential for expressivity in this task; the measured observable remains a linear function of the encoded inputs regardless of where mode mixing occurs. In contrast, relocating the rotation gates to the end of the circuit reduced accuracy to approximately $92.0\%$ (base accuracy $97.33\%$). This modification alters how parameter perturbations influence the measured quadrature, effectively distorting the sensitivity directions encoded in the Jacobian.

Introducing squeezing operations at the beginning of the circuit further degraded performance to approximately $81.3\%$. Squeezing introduces nonlinearities and so reshapes the circuit’s response manifold and can lead to a poorly conditioned Jacobian, making the local inverse problem harder to solve using linearized pseudoinverse updates. Similarly, fixing the beam splitter parameters reduced accuracy to approximately $78.7\%$, demonstrating that mode-mixing flexibility is crucial for spanning independent sensitivity directions in parameter space. Reintroducing trainable displacement scaling partially restored performance to approximately $92.0\%$, showing that increasing the space of modifiable parameters can recover controllability of the output manifold.

In sum, these architectural perturbations demonstrate that learning success depends primarily on how circuit structure shapes the conditioning and expressivity of the Jacobian. Operations that preserve a well-conditioned mapping between parameter perturbations and measurable outputs enable stable Gauss–Newton–like corrections, whereas operations that distort or restrict this mapping degrade learning. This underscores that PCPL performance is governed by response geometry rather than circuit complexity alone.

\subsubsection{Tunable linear optical multiplier: Towards an Autonomous Physical Learner}
Mode~B PCPL requires a local update signal proportional to the product of a sensitivity and an error. In classical gradient descent learning, this appears as
\begin{equation}
\Delta \theta \propto \left(\frac{\partial f}{\partial \theta}\right)\bigl(z_{\text{target}} - z\bigr),
\end{equation}

so any fully physical implementation of PCPL must be able to generate such multiplicative signals from measurable contrasts. Here we address the complementary requirement: whether the multiplication of that sensitivity by an error signal can itself be realized within a physical optical system.

The mechanism we exploit—interference between displacement-encoded optical modes followed by quadrature measurement—relies only on linear optical elements and coherent states. As a result, an equivalent multiplier could in principle be implemented using purely classical electromagnetic fields. Interference of classical optical amplitudes and square-law detection have long been used in analog optical processors to perform multiplicative and convolution-like operations. The present construction does not claim novelty at the level of optical physics; rather, its significance lies in embedding such interference-based multiplication directly into the structure of a PCPL learning rule. In this sense, the circuit functions as a {\it physical gradient unit}, translating contrast measurements into update signals within the same substrate that performs inference. This represents a step toward partially autonomous physical learners, in which not only inference but also elements of the update rule are implemented directly in the physical substrate. We define a physical learning system as autonomous to the extent that components of the parameter-update rule are implemented within the same physical substrate that performs inference, rather than being computed by an external digital processor.

\noindent{\bf Goal.} Our goal is to construct a linear optical circuit whose measured output approximates the analog product $\delta f \cdot e$, thereby embedding a fundamental part of the learning operation into the measurement process. We consider two approaches within linear optics. In the first approach, the signals $\delta f$ and $e$ are encoded into quadrature amplitudes via displacement gates and the encoded modes are coupled through a beam splitter. A linear fit is then applied to $x$-quadrature mean measurements outcomes to find parameters 
that minimize the approximation error. Note this gives you $\delta f\pm e$, ie, $\delta f$ and $e$ separately, not their product.

In the second approach, we exploit the intensity-difference property of a beam splitter. Encoding $\delta f$ and $e$ into the two input modes, the beam splitter produces output amplitudes $\alpha_0 = (\delta f - e)/\sqrt{2}$ and $\alpha_1 = (\delta f + e)/\sqrt{2}$. The difference of squared direct detection then yields
\begin{equation}
\frac{\delta f}{\varepsilon} \cdot e=\frac{\alpha_1^2 - \alpha_0^2}{2\cdot\varepsilon} \approx g_k,
\end{equation}
where $g_k$ is the gradient estimate weighted by the error, obtained from intensity measurements or classical post-processing of $x$-quadrature outputs, scaled by the fixed perturbation size $\varepsilon$, so the product $(\delta f / \varepsilon) \cdot e$ can be recovered directly from the intensity measurement without requiring any nonlinear optical elements. Both approaches yield comparable final classification accuracy.

\begin{figure}[]
\centering	
\includegraphics[width=0.5\textwidth]{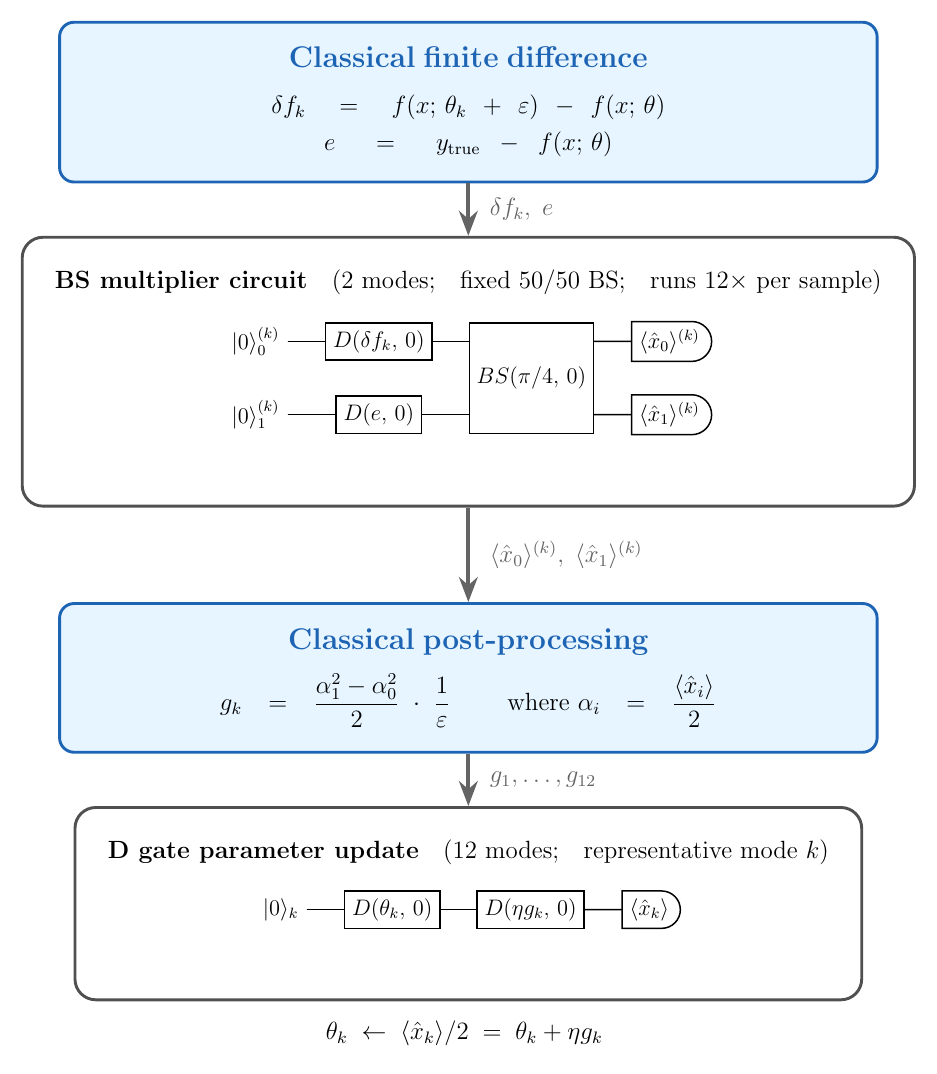}
\caption{\textit{Simplest linear optical learning circuit for Iris dataset classification.} The circuit consists of a classical finite-difference block, a two-mode CV BS multiplier circuit for gradient computation (run once per parameter per sample), a classical post-processing block for intensity-difference readout, and a 12-mode D gate parameter update circuit with x-quadrature mean readout.}
\label{fig:linear_optical_multiplier}
\end{figure}

\noindent{\bf Implementation.} The parameter update is implemented by a second CV circuit in which each parameter mode is initialized via a displacement $D(\theta_k)$ and the scaled gradient is added by a second displacement $D(\eta g_k)$, so that the updated value is read from the quadrature mean $\langle \hat{x}_k \rangle / 2 = \theta_k + \eta g_k$. This minimal circuit thus represents a CV self-learning system in which gradient computation and parameter update all occur within integrated Gaussian circuits, with the exception of classical finite-difference perturbations used to evaluate $\delta f$ and parameter clipping applied to the updates. See Fig. \ref{fig:linear_optical_multiplier}.

We now incorporate the linear optical multiplier into the larger linear optical learning circuit. To classify the Iris dataset, the system uses a 4-mode photonic classifier with 12 trainable parameters (4 rotation angles and 8 beam splitter parameters). The classifier encodes iris features through displacement gates $\hat{D}(x_i)$, applies learnable rotations $\hat{R}(\theta_i)$, and processes the state through two layers of parameterized beam splitters followed by an additional adaptive rotation layer and two final beam splitter operations. The forward pass produces a continuous $x$-quadrature mean measurement output.

\noindent{\bf Performance.} The minimal two-mode linear optical multiplier performs parameter updates through a dedicated two-mode displacement circuit: each parameter mode is initialized by $\hat{D}(\theta_k)$ and the scaled gradient is added by $\hat{D}(\eta g_k)$, so the updated value is recovered as $\langle \hat{x}_k \rangle / 2 = \theta_k + \eta g_k$. No additional processing stages are involved. The system achieves $96.2\%\pm 2.3\%$ (Configuration A, 20 epochs)/training and testing accuracy for the Iris dataset (25 independent realizations), demonstrating that photonic physical self-learning with photonic-computed gradients is feasible. See Appendix C for an example with squeezing as well as the effect of circuit architecture on the performance.

\section{Discussion}

This work introduces Perturbative Contrastive Physical Learning (PCPL) as a unifying principle for learning in physical systems. Rather than defining learning through explicit gradients of a loss function, PCPL frames learning as the measurable contrast between nearby physical states produced by controlled perturbations. In this view, the fundamental primitive of learning is not symbolic differentiation but experimentally accessible response. By elevating contrast to the central role, PCPL provides a substrate-independent description of learning that applies equally to mechanical, optical, and photonic systems. This reframing shifts the question from “How does one compute gradients?” to “What contrasts can a physical system sense and translate into parameter change?”

A key insight of the PCPL framework is the distinction between two learning geometries. Mode A learning probes local sensitivity: the system compares its response before and after a small internal perturbation, effectively measuring a Jacobian action. Mode B learning performs local inversion: contrasts between nudged responses and targets induce Gauss–Newton–like parameter updates governed by the natural metric of the system’s input–output map. Importantly, this inverse geometry need not be computed explicitly. In the physical implementations shown here, the effective metric emerges implicitly from the system’s response, demonstrating that metric-aware learning can be realized through local perturbation and measurement rather than global matrix inversion. In the two platforms, a spring network and a CV photonic gates based circuit, learning is successful, suggesting that learning performance in physical substrates depends less on whether a system is classical or photonic, and more on how its architecture shapes the conditioning and expressivity of its response manifold.

The tunable continuous-variable photonic multiplier extends this idea by embedding part of the learning rule itself as part of the physical substrate. By implementing an analog multiplication between a sensitivity signal and an error signal through optical interference, we show that a core computational primitive of Gauss–Newton learning can migrate from classical post-processing into the physical substrate. This represents an initial step toward autonomous physical learners in which inference and elements of parameter update emerge from the same substrate. In such systems, learning is no longer an algorithm imposed from outside, but an algorithmic process shaped by the system’s own response geometry.

Previous work has demonstrated that physical systems can be trained through response-based adaptation, including the programming of stress and strain patterns in spring networks and cell packings in disordered materials~\cite{Hexner2023,ameen2026training}. These approaches emphasize experimentally realizable update rules, often with a strong focus on strictly local adaptation. PCPL does not adopt locality as a defining principle. Instead, it is organized around the geometry of physically measurable contrasts and the contrast measurement functional $M$ need not act on
strictly local observables. In fact, the measurable contrasts may involve
collective, nonlocal, interferometric, or globally encoded response
variables, provided that they contain sufficient information about the
underlying response geometry of the system. In Mode A, these contrasts probe local sensitivities, while in Mode B they can encode information about a global inverse problem through the system’s collective response. Importantly, this does not imply centralized learning: no external processor computes gradients or inverts system-wide Jacobians. Instead, the learning signals arise directly from the system’s own physical response to controlled perturbations. While the resulting updates can reflect global properties of the system, such as collective modes, long-range couplings, or inverse-problem geometry, these effects are mediated through the substrate’s intrinsic response rather than through an explicit global optimization algorithm. In other words, the system does not require a separate computational layer with access to a full model of its state; the effective learning geometry emerges implicitly from measurable contrasts between nearby physical configurations. In PCPL, by contrast, the “computation” of update directions is distributed across the physical degrees of freedom themselves, and is realized through experimentally accessible perturbation–response measurements rather than algorithmic inversion. PCPL therefore occupies an intermediate regime between strictly local adaptation and centrally computed gradient descent, demonstrating that nonlocal learning behavior can arise without centralized algorithmic control. 

As for experimental implications, the optical character of CV circuit brings about the distinct possibility of physical learning with low energy consumption compared to Joule-heating electronic circuits, even if one must take into account the amount of laser power necessary to drive the circuit and, specifically, to pump the optical parametric oscillators that produce squeezed light. Nevertheless, perspectives are enticing, especially in light of the constantly evolving state of the art of integrated photonics~\cite{Bogaerts2020}, as integration begets scalability. Learning with continuous quantum variables using Gaussian and non-Gaussian gates~\cite{Braunstein2005a} has been studied extensively, including Continuous-Variable Quantum Neural Networks~\cite{Killoran2019} and relevant applications to state classification (with classical feedforward post-processing)~\cite{shokou2025} and time-series forecasting (with classical optimization)~\cite{Anand2024}. In addition, recent work has demonstrated that learning can be implemented in linear optical interferometers with coherent states and homodyne measurements~\cite{anteneh2026laser}. These studies demonstrate that CV quantum circuits can be used for learning tasks; however, classical computational processing blocks are still required. In other words, to realize a fully autonomous quantum learning system, aka a quantum creature, perturbations, contrasts using single-shot measurements, multiplication/error couplings, and parameter updates are all implemented within the quantum substrate itself. We are currently working towards this goal.

More broadly, PCPL expands the notion of what constitutes a learnable parameter. In physical systems, modifiable degrees of freedom include not only coupling strengths but also geometry, boundary conditions, and even interpretive mappings between physical states and functional outputs. By defining learnability operationally—through what can be perturbed and sensed—PCPL shifts the focus from abstract optimization to experimentally accessible contrasts between physical states. This perspective opens pathways for designing adaptive materials, programmable photonic circuits, hybrid classical–quantum, and fully quantum, platforms, where learning is embedded directly in physical laws.

\noindent{\it Acknowledgements.} We thank Benjamin Scellier for useful discussions. JMS acknowledges financial support from the National Science Foundation via DMR-2204312.  OP was supported by National Science Foundation grants OSI-2531569 [NQVL:QSTD:Design: Quantum Computing Applications of Photonics (QCAP)], PHY-2514971, and ECCS-2530171.

\setlength{\emergencystretch}{2em}
\bibliographystyle{unsrt}
\bibliography{references,Pfister}

\section*{Appendix A: Mode~B Gradient Descent vs. Pseudoinverse Performance Comparison}

\begin{figure*}[]
    \centering
    \includegraphics[width=14cm]{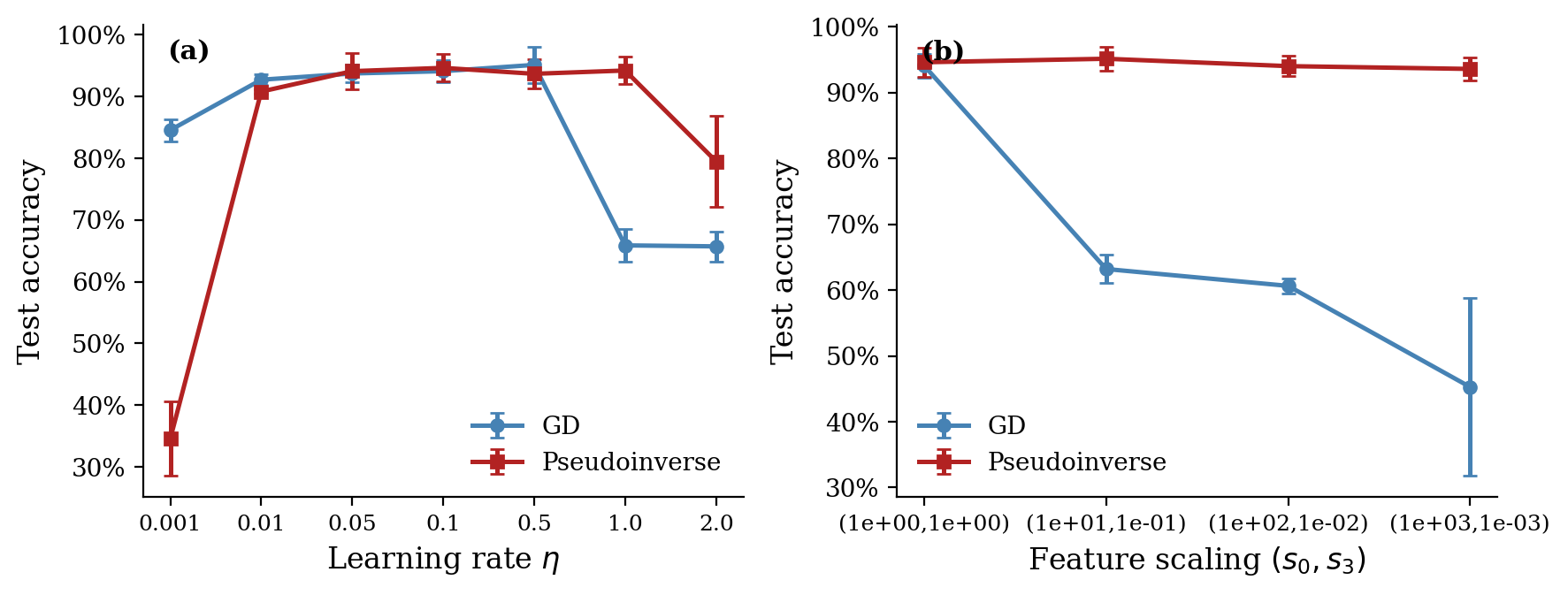}  
    \caption{\textit{Robustness comparison between gradient descent (GD) and pseudoinverse parameter updates for Iris classification.} (a) Test accuracy as a function of learning rate $\eta$. GD maintains competitive performance for small $\eta$ but degrades sharply above $\eta = 0.5$, whereas the pseudoinverse update remains stable across the full range tested. (b) Test accuracy under increasingly ill-conditioned feature scaling $(s_0, s_3)$, where $s_0$ and $s_3$ denote the scaling factors applied to the first and last features, respectively. GD accuracy drops substantially as the condition number increases, while the pseudoinverse update retains above $93\%$ accuracy throughout. Error bars indicate one standard deviation over 25 independent trials.}
    \label{fig:GD_vs_pseudoinverse}
\end{figure*}

We compare gradient descent (GD) and the pseudoinverse update rule on a linear classifier trained on the Iris dataset (50\%/50\% train--test split, 25 independent trials). Two experiments are conducted to isolate the effect of the update rule (Fig. \ref{fig:GD_vs_pseudoinverse}). First, we sweep the learning rate $\eta$ over a wide range to test sensitivity: GD requires careful tuning of $\eta$, whereas the pseudoinverse update is self-normalizing via $(J^\top J)^{-1}$ and therefore robust to the choice of $\eta$. Second, we apply uneven feature scaling to introduce ill-conditioning: one feature is amplified by a factor of $10^2$ (or $10^3$) while another is suppressed by $10^{-2}$ (or $10^{-3}$), creating a highly anisotropic loss landscape in which GD oscillates or diverges while the pseudoinverse update converges stably. In both experiments, adaptive thresholds and parameter clipping are deliberately excluded to isolate the effect of the update rule alone; these techniques can improve accuracy by up to $4\%$ in practice. Both methods use a 1-output linear model with fixed thresholds.

Table~\ref{tab:method_comparison} demonstrates this on a linear classifier trained on the Iris dataset: while GD and the pseudoinverse update achieve comparable accuracy at a well-tuned learning rate ($\eta=0.05$--$0.1$), GD degrades rapidly for larger $\eta$ or under feature ill-conditioning, whereas the pseudoinverse update maintains consistent performance throughout.
\begin{table*}[]
\centering
\begin{tabular}{l|c|c|c|c}
\hline
Method & $\eta=0.05$ & $\eta=1.0$ & scaling $(10^2, 10^{-2})$ & scaling $(10^3, 10^{-3})$ \\
\hline
GD & $93.7\%\pm1.4\%$ & $65.9\%\pm2.7\%$ & $60.6\%\pm1.1\%$ & $45.3\%\pm13.5\%$ \\
Pseudoinverse & $94.1\%\pm2.9\%$ & $94.2\%\pm2.2\%$ & $94.0\%\pm1.6\%$ & $93.6\%\pm1.8\%$ \\
\hline
\end{tabular}
\caption{\textit{Test accuracy for Iris classification(50\%/50\% split, 25 trials) comparing GD and pseudoinverse update under varying learning rates and feature ill-conditioning.} Note that additional techniques such as adaptive thresholds and parameter clipping were deliberately excluded to isolate the effect of the update rule; these can improve accuracy by up to $4\%$ in practice.}
\label{tab:method_comparison}
\end{table*}
The numerical results in Table~\ref{tab:method_comparison} confirm that the pseudoinverse update of Mode~B is robust to hyperparameter choice and input ill-conditioning, a consequence of its metric-aware correction in the natural geometry of the system's response manifold.

\section*{Appendix B: More details on PCPL}
\subsection*{Least Squares Problem with Tikhonov regularization}
For $N$ input samples, we collect the measured quadrature outputs into a vector $\mathbf{z}\in \mathbb{R}^N$ and define $\mathbf{z}^*$ as the desired class-dependent targets. Assuming linear response of the system with respect to parameters $\bm{\theta} \in \mathbb{R}^p$, the residual $\mathbf{e} = \mathbf{z}^* - \mathbf{z}$ satisfies
\begin{equation}
\mathbf{e} \approx J\,\Delta\bm{\theta},
\end{equation}
where $J \in \mathbb{R}^{N \times p}$ is the Jacobian of the output observable with respect to the system parameters. We seek the parameter update $\Delta\bm{\theta}$ that minimizes the regularized objective
\begin{equation}\label{LSQeq1}
\min_{\Delta\bm{\theta}}\;
\bigl\|J\,\Delta\bm{\theta} - \mathbf{e}\bigr\|^2
+ \lambda\,\|\Delta\bm{\theta}\|^2,
\end{equation}
where the second term $\lambda\,\|\Delta\bm{\theta}\|^2$ penalizes large parameter updates and $\lambda > 0$ controls the strength of regularization. Taking the derivative with respect to $\Delta\bm{\theta}$ and setting it to zero yields
\begin{align}
&\frac{\partial}{\partial\,\Delta\bm{\theta}}
\Bigl[\bigl\|J\,\Delta\bm{\theta} - \mathbf{e}\bigr\|^2
+ \lambda\,\|\Delta\bm{\theta}\|^2\Bigr]\notag\\
&= 2J^\top(J\,\Delta\bm{\theta} - \mathbf{e})
   + 2\lambda\,\Delta\bm{\theta} = \mathbf{0},
\end{align}
which gives $(J^\top J + \lambda I)\,\Delta\bm{\theta} = J^\top\mathbf{e}$. The closed-form solution is thus the regularized pseudoinverse step
\begin{equation}\label{LSQeq2}
\Delta\bm{\theta}
= \bigl(J^\top J + \lambda I\bigr)^{-1} J^\top \mathbf{e},
\qquad \lambda > 0
\end{equation}
to arrive at Eq. 10. 

\subsection*{Linear Response and the Contrast Measurement Functional $M$ }

To obtain a direct mapping from parameter perturbations and $M$, we start with the approximation 
\begin{equation}
G(\mathbf{y}, \bm{\theta}') - G(\mathbf{y}, \bm{\theta}) \approx \sum_i \frac{\partial G}{\partial \theta_i}(\mathbf{y}, \bm{\theta}) \cdot (\theta'_i - \theta_i).
\end{equation}
The RHS can be written more compactly using the Jacobian matrix $J_{\bm{\theta}} G$ of partial derivatives:
\begin{equation}
G(\mathbf{y}, \bm{\theta}') - G(\mathbf{y}, \bm{\theta}) \approx J_{\bm{\theta}} G(\mathbf{y}, \bm{\theta}) \cdot (\bm{\theta}' - \bm{\theta}),
\end{equation}\label{eq-G-1storder-approx}
where $J_{\bm{\theta}} G(\mathbf{y}, \bm{\theta})$ is the Jacobian matrix with elements:
\begin{equation}
[J_{\bm{\theta}} G(\mathbf{y}, \bm{\theta})]_{j,i} = \frac{\partial G_j}{\partial \theta_i}(\mathbf{y}, \bm{\theta}),
\end{equation}
and $G_j$ is the $j$-th component of the output vector $G$.

Similarly, the contrast measurement functional $M$ applied to the output contrast can be approximated using its own Jacobian:
\begin{align}
&M(G(\mathbf{y}, \bm{\theta}'), G(\mathbf{y}, \bm{\theta})) \notag\\
&\approx J_{G} M(G,G') \cdot (G(\mathbf{y},\bm{\theta}') - G(\mathbf{y},\bm{\theta})),    
\end{align}
where $G' = G(\mathbf{y},\bm{\theta}')$ and $J_{G} M(G,G')$ is the Jacobian of the measurement operator with respect to changes in $G$. Combining these two approximations, we can derive the relationship between parameter perturbations and measurement outputs:
\begin{align}
&M(G(\mathbf{y}, \bm{\theta}'), G(\mathbf{y}, \bm{\theta})) \notag\\&\approx J_{G} M(G,G') \cdot (G(\mathbf{y},\bm{\theta}') - G(\mathbf{y},\bm{\theta})), \notag\\
&\approx J_{G} M(G,G') \cdot J_{\bm{\theta}} G(\mathbf{y}, \bm{\theta}) \cdot (\bm{\theta}' - \bm{\theta}).
\end{align}
This gives us a direct mapping from parameter perturbations to measurement outputs through the composition of two Jacobian transformations. 

\subsection*{Gradient Descent Approximation and Beyond}
We start with contrast measurement functionals that satisfy:
\begin{equation}
J_{G} M(G,G') \cdot J_{\bm{\theta}} G(\mathbf{y}, \bm{\theta}) \approx \gamma \cdot \nabla_{\bm{\theta}} \mathcal{L}(\bm{\theta})^T,
\end{equation}
for some positive constant $\gamma$ and loss function $\mathcal{L}(\bm{\theta})$. One way to achieve this is to design the measurement operator such that:
\begin{equation}
J_{G} M(G,G') \approx \lambda \cdot (J_{\bm{\theta}} G(\mathbf{y}, \bm{\theta}))^+,
\end{equation}
where $(J_{\bm{\theta}} G)^+$ is the pseudoinverse of the Jacobian and $\lambda$ is a scaling factor. This design choice enables us to construct a parameter update rule that approximates gradient descent. With this design, the parameter update becomes:
\begin{align}
\bm{\theta}_{\text{new}} &= \bm{\theta} + \eta \cdot M(G(\mathbf{y}, \bm{\theta}'), G(\mathbf{y}, \bm{\theta})) \\
&\stackrel{(i)}{\approx} \bm{\theta} + \eta \cdot J_{G} M(G,G') \cdot J_{\bm{\theta}} G(\mathbf{y}, \bm{\theta}) \cdot (\bm{\theta}' - \bm{\theta}) \\
&\stackrel{(ii)}{\approx} \bm{\theta} + \eta\lambda \cdot (J_{\bm{\theta}} G)^+ \cdot J_{\bm{\theta}} G \cdot (\bm{\theta}' - \bm{\theta}) \\
&\stackrel{(iii)}{=} \bm{\theta} + \eta\lambda \cdot (\bm{\theta}' - \bm{\theta}),
\end{align}
where (i) applies the first-order linear approximation, (ii) substitutes the design from equation (2), and (iii) uses the pseudoinverse property $(J_{\bm{\theta}} G)^+ \cdot J_{\bm{\theta}} G \approx I$.

For nudging rules where $\bm{\theta}' - \bm{\theta} \propto \nabla_{\bm{\theta}} \mathcal{L}(\bm{\theta})$, this results in effective gradient descent toward minimizing the loss function. However, more generally one can treat $G(\bm{y},\bm{\theta})$ as a black box and measure linear response to update $\bm{\theta}$ given the infinitesimal linear approximation. This approach works well for single-output problems and small multi-output cases ($n \leq 3$). For higher-dimensional outputs ($n > 3$), on the other hand, the inverse mapping $\delta\bm{\theta} \propto G^{-1}(\bm{z}_{\text{target}} - G(\bm{y},\bm{\theta}))$ relies heavily on infinitesimal linear approximations through Jacobian-based updates. In such cases, establishing explicit non-linear models for the intermediate $y \rightarrow z$ mappings will presumably improve training efficiency compared to relying solely on local linear approximations.

The accuracy of such approximations depends on the size of the perturbation $\|\bm{\theta}' - \bm{\theta}\|$ and the nonlinearity of $G$. The error term is:
\begin{equation}
\text{Error} = G(\mathbf{y}, \bm{\theta}') - G(\mathbf{y}, \bm{\theta}) - J_{\bm{\theta}} G(\mathbf{y}, \bm{\theta}) \cdot (\bm{\theta}' - \bm{\theta}).
\end{equation}
For sufficiently small perturbations, this error becomes negligible compared to the first-order term, which is the key assumption that enables effective parameter learning through PCPL.

\section*{Appendix C: More on tunable multipliers}
\subsection*{Tunable multiplier with squeezing}
\begin{figure*}
    \centering
    \begin{tabular}{c}
\begin{tikzpicture}[
    >=Stealth,
    line width=0.7pt,
    every node/.style={font=\small},
    scale=1, transform shape,
]
\node (circ) {
    \begin{quantikz}[column sep=0.55cm, row sep=0.44cm]
        \lstick{$|0\rangle$}
        & \gate{D(d_\ell \cdot e,\,0)}
        & \gate{S(s_\ell \cdot \delta f)}
        & \gate{R(\phi_\ell)}
        & \meterD{\langle\hat{x}_0\rangle} \\
    \end{quantikz}
};
\end{tikzpicture}\\
(a) Trainable circuit diagram\\
\\
    \includegraphics[width=0.8\linewidth]{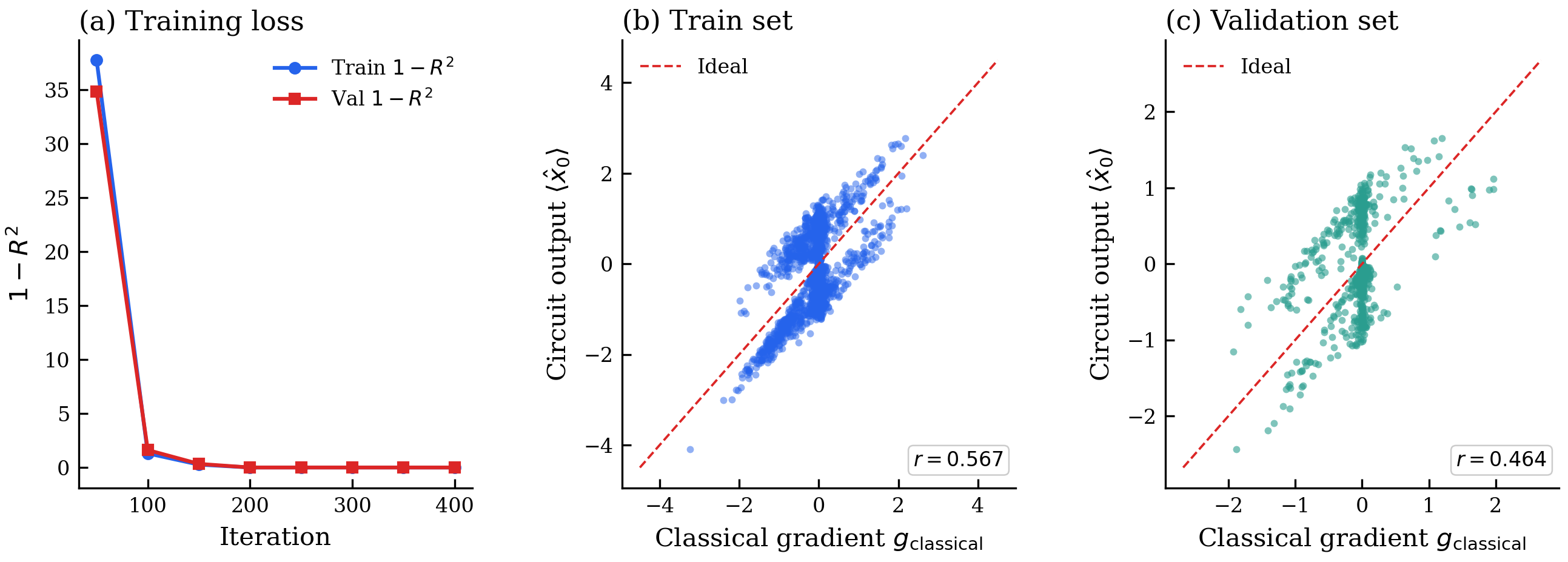}\\
    \\
    (b) Training performance\\
    \end{tabular}
    \caption{\textit{DSR multiplier.}}
    \label{fig:DSR_multiplier}
\end{figure*}

We consider a tunable multiplier using squeezing gates. In a single-mode Gaussian circuit, applying a displacement $\hat{D}(d\cdot e)$ followed by squeezing $\hat{S}(s\cdot\delta f)$ to vacuum gives an $x$-quadrature mean that depends on both signals. Here $d$ and $s$ are trainable parameters, so $d\cdot e$ and $s\cdot\delta f$ are simply classical signals scaled by fixed gains before encoding into the circuit. However, we will show how to find suitable values based on the target system. The interaction between the two signals arises from the nonlinear dependence $e^{-s\cdot\delta f}$ introduced by the squeezing gate acting on the already-displaced state.
\begin{equation}
    \langle \hat{x} \rangle = \sqrt{2}\, d \cdot e \cdot e^{-s\,\delta f}
\end{equation}
since the squeezing scales the displacement mean by $e^{-s\,\delta f}$, where rotation is not yet applied. In the regime $s\cdot\delta f \ll 1$ this approximates:
\begin{equation}
    \langle \hat{x} \rangle \approx \sqrt{2}\, d \cdot e \cdot (1 - s\,\delta f)
\end{equation}
which contains the product term $s \cdot d \cdot \delta f \cdot e$ and the first term is fixed values for fixed $e$. We therefore define a single-mode circuit layer as:
\begin{equation}
    \hat{R}(\phi)\,\hat{S}(s \cdot \delta f)\,\hat{D}(d \cdot e)\,|0\rangle
\end{equation}

where $s$, $d$, and $\phi$ are trainable parameters and the rotation $\hat{R}(\phi)$ aligns the output quadrature with the measurement axis, giving $\langle\hat{x}\rangle = \sqrt{2}\,d\cdot e\cdot e^{-s\,\delta f}\cos\phi$. This circuit is trained to minimize $1 - R^2$ between its output and the classical gradient $(\delta f / \varepsilon) \cdot e$ over a dataset of realistic $(\delta f, e)$ pairs generated from the CV linear classifier with random parameters. We refer to this as the \emph{DSR circuit}. The full circuit with $L$ layers has trainable parameters $\{s_\ell, d_\ell, \phi_\ell\}_{\ell=1}^{L}$, totalling $3L$ parameters, with output $\langle \hat{x}_0 \rangle$. Figure~\ref{fig:DSR_multiplier} shows the circuit diagram and the training performance. 

The inputs $\delta f = f(x;\,\theta_i + \varepsilon) - f(x;\,\theta)$ and $e = y_{\mathrm{true}} - f(x;\,\theta)$ are computed classically before being encoded into the circuit. The circuit parameters $\{s_\ell, d_\ell, \phi_\ell\}_{\ell=1}^{L}$ are pre-trained offline and fixed during Iris classifier training.
Specifically, the DSR multiplier is trained offline using a dataset of $(\delta f, e, g_{\mathrm{classical}})$ tuples generated by running the CV linear classifier with randomly sampled parameter vectors $\bm{\theta} \sim \mathrm{Uniform}(-0.3, 0.3)^{12}$ on Iris training samples, with $n_{\theta} = 50$ random parameter sweeps per sample. For each sweep, a random parameter index $i$ is selected, the finite difference $\delta f = f(x;\,\theta_i + \varepsilon) - f(x;\,\theta)$ is computed with $\varepsilon = 0.015$, and the error $e = y_{\mathrm{true}} - f(x;\,\theta)$ is recorded. The circuit parameters are optimized by minimizing $1 - R^2$ between the circuit $x$-quadrature mean output and $g_{\mathrm{classical}} = (\delta f / \varepsilon) \cdot e$ using a coordinate-wise random search optimizer over 400 iterations with initial learning rate $0.1$ and decay factor $0.997$. Inputs $\delta f$ and $e$ are normalized by their training-set standard deviations before encoding.

\begin{figure*}
    \centering
    \includegraphics[width=0.6\linewidth]{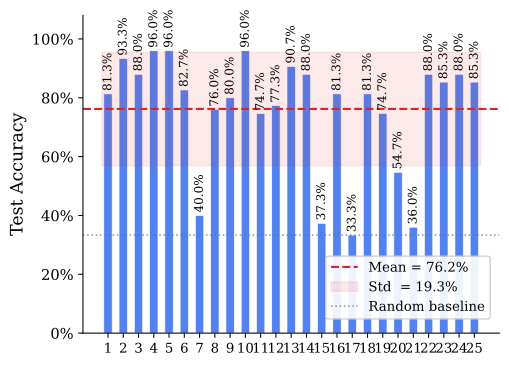}
    \caption{\textit{Performance of the DSR multiplier across trials.}}
    \label{fig:DSRMultiplier}
\end{figure*}

Now we test for classification by combining the DSR multiplier (with pre-trained data) and the D gate parameter update circuit (same as the main text). Figure~\ref{fig:DSRMultiplier} shows the classification performance with this combined circuit. Learning is inconsistent across trials due to approximation error introduced by the nonlinear squeezing-based multiplication, which only approximates $\delta f \cdot e$ in the small-signal regime $s \cdot \delta f \ll 1$. When this condition is violated, the gradient signal deviates from the true error-weighted finite difference, causing parameter updates to point in incorrect directions. Within individual trials, fluctuations in training accuracy are also observed; to mitigate this, we record the best-performing parameters during training and restore them at the end (15 epochs, learning rate $\eta = 1.0$). Nevertheless, in successful trials the best accuracy reaches $96.0\%$, demonstrating that the DSR multiplier can provide sufficient gradient information for effective learning when the operating conditions are favorable. The inconsistency arises from two main sources: (1) parameter range clipping in the D gate update circuit, which truncates large parameter updates and prevents the classifier from reaching certain regions of parameter space; and (2) the nonlinearity of the squeezing-based multiplication, which introduces a systematic bias in the gradient estimate that grows with the magnitude of $\delta f$. Since both sources of error are data- and initialization-dependent, some random seeds lead to favorable operating regimes where the approximation holds well throughout training, while others do not. This sensitivity to initialization is an inherent limitation of the pure CV approach without planned circuit design. A more carefully designed CV multiplier circuit, with explicit consideration of the operating range of $\delta f$ and $e$ and a circuit architecture that reduces the dependence on the small-signal approximation, would be expected to improve both consistency and accuracy across trials.

\subsection*{Tunable linear optical multiplier}: Configuration A (from main text) and classical update comparison

\begin{figure*}[]
\centering
\begin{tabular}{c}
      \includegraphics[width=14cm]{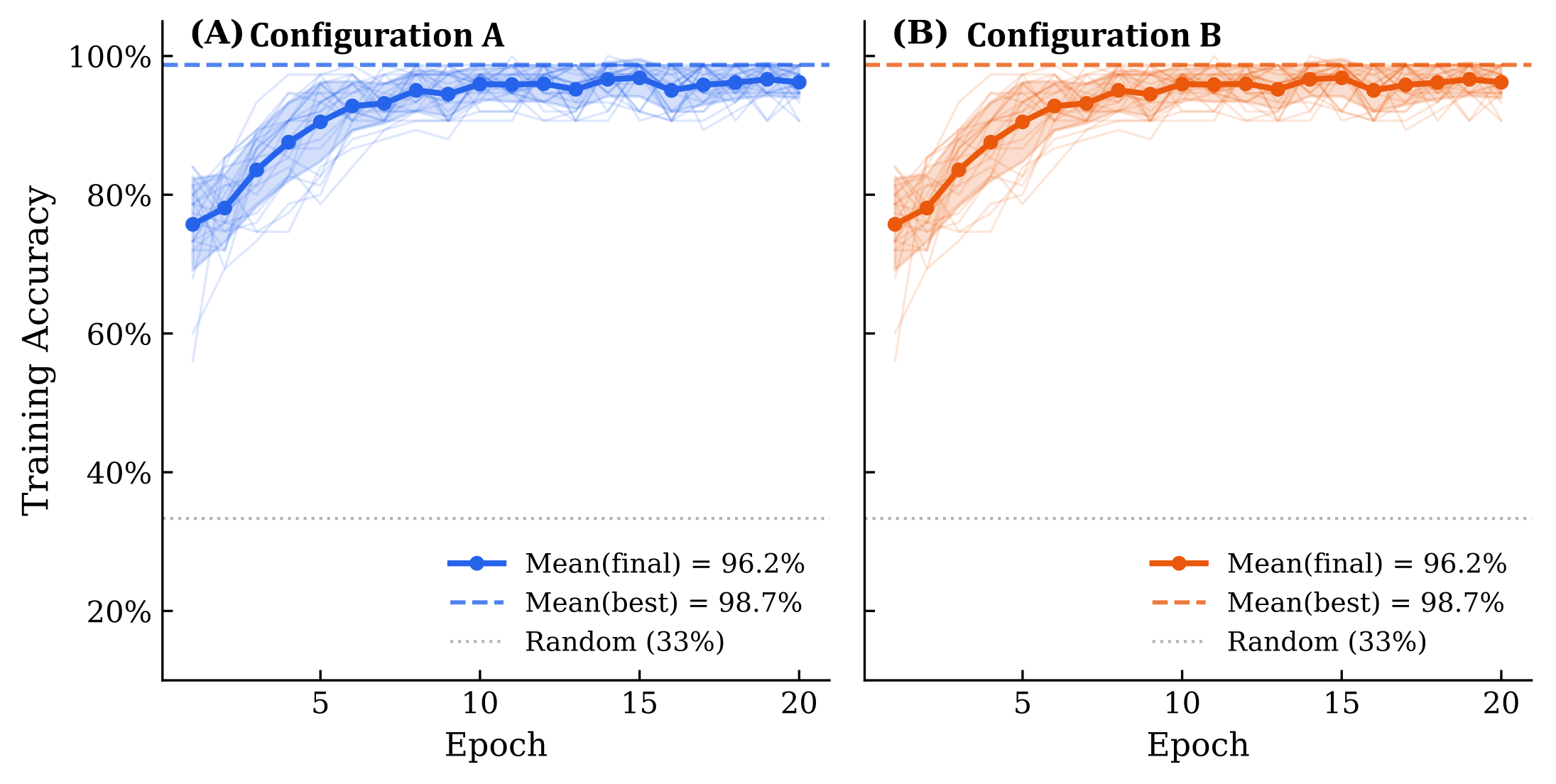}  
      \\
      \includegraphics[width=14cm]{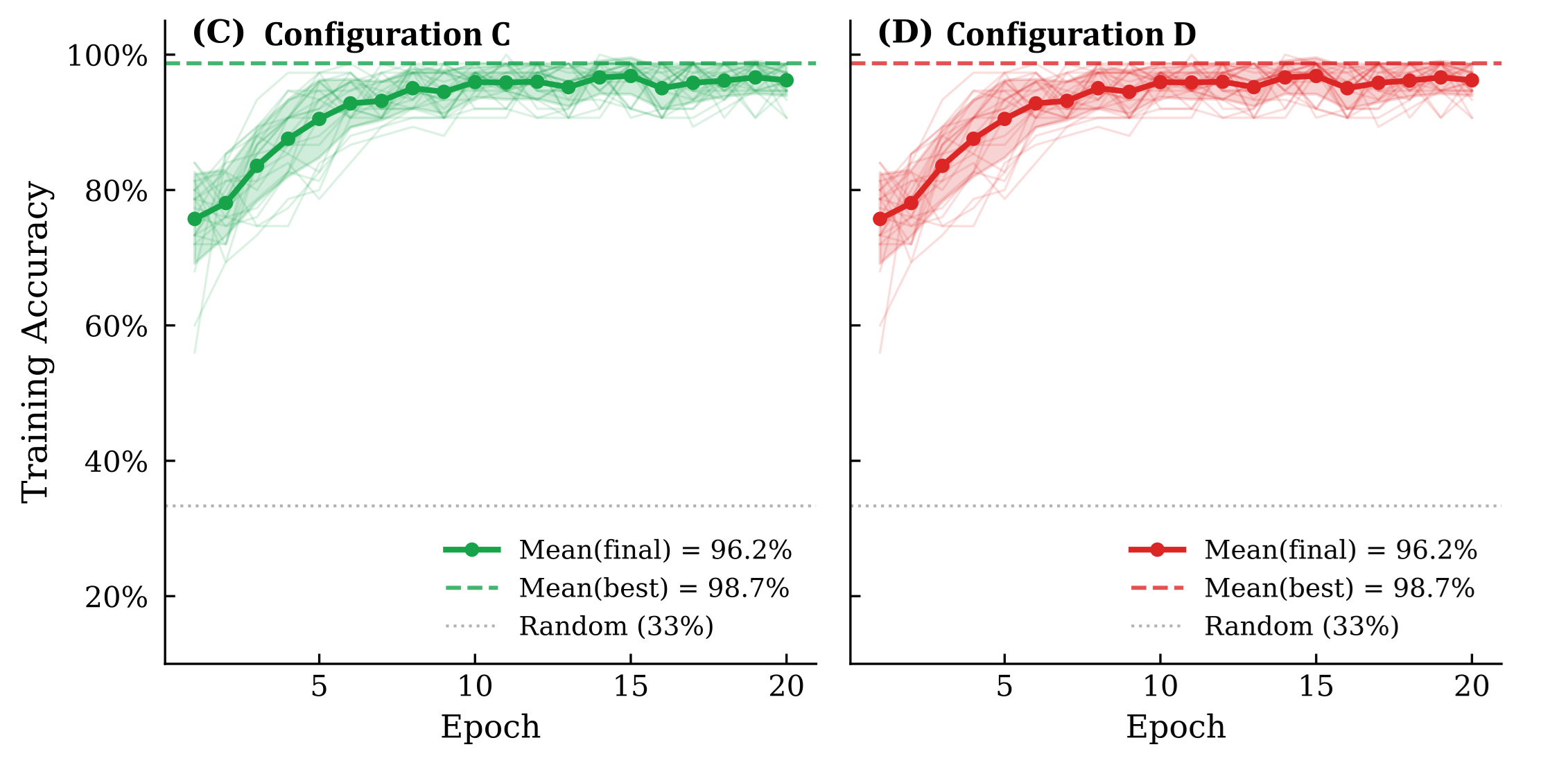}  
\end{tabular}
\caption{\textit{Performance comparison among different combinations of classical and CV learning components using the same seeds.}}
\label{fig:performance_comparison}
\end{figure*}

To isolate the contribution of each CV component, we compare four configurations: 
\begin{itemize}
    \item Configuration~A pairs the linear optical multiplier gradient with the CV displacement update (the full CV system)
    \item Configuration~B replaces the CV gradient with a classical finite-difference gradient while retaining the CV displacement update
    \item Configuration~C retains the CV gradient but replaces the displacement update with classical gradient descent
    \item Configuration~D uses classical gradients and 
classical gradient descent throughout, serving as the full classical baseline.
\end{itemize}
Table~\ref{tab:ablation} summarizes the results across 25 independent trials. 
All four configurations achieve identical performance: $98.7\%$ best test accuracy, 
$96.2\%$ final test accuracy, and $2.3\%$ standard deviation. The absence of any 
measurable difference across configurations indicates that the linear optical multiplier gradient 
is functionally equivalent to the classical finite-difference gradient, and that the 
CV displacement update is functionally equivalent to classical gradient descent, 
at least for this task. This is consistent with the theoretical expectation that 
the linear optical multiplier circuit computes $\delta f \cdot e / \varepsilon$ to machine precision 
via the beam splitter intensity-difference identity, and that the displacement update 
circuit implements arithmetic addition exactly within the Gaussian backend. 
The CV components therefore do not introduce approximation errors relative to their 
classical counterparts, and the full CV system of Configuration~A achieves 
classical-equivalent accuracy while realizing both gradient computation and 
parameter update as physical analog operations. See Fig. \ref{fig:performance_comparison}. 
\begin{table*}[]
\centering
\caption{Ablation results across 25 independent trials (20 epochs).}
\label{tab:ablation}
\begin{tabular}{lrrrrr}
\hline
Config & Train (best) & Train (final) & Test (best) & Test (final) & Std (test) \\
\hline
A & 98.7\% & 96.3\% & 98.7\% & 96.2\% & 2.3\% \\
B & 98.7\% & 96.3\% & 98.7\% & 96.2\% & 2.3\% \\
C & 98.7\% & 96.3\% & 98.7\% & 96.2\% & 2.3\% \\
D & 98.7\% & 96.3\% & 98.7\% & 96.2\% & 2.3\% \\
\hline
\end{tabular}
\end{table*}

\end{document}